\documentclass{article}

\usepackage{arxiv}

\usepackage{microtype}
\usepackage{graphicx}
\usepackage{subcaption}
\usepackage{booktabs} 

\usepackage{hyperref}

\usepackage{amsmath}
\usepackage{amssymb}
\usepackage{mathtools}
\usepackage{amsthm}

\usepackage[capitalize,noabbrev]{cleveref}

\theoremstyle{plain}

\theoremstyle{definition}

\theoremstyle{remark}

\usepackage{natbib}
\usepackage{tikz}
\usepackage{tikz-cd}
\usepackage{booktabs}

\RequirePackage{doi}
\usepackage{hyperref}
\usepackage{pgfplots}
\usepackage{subcaption}
\usetikzlibrary{arrows.meta,calc,positioning,decorations.pathreplacing}
 \pgfplotsset{compat=1.18}

\usepackage{algorithm}
\usepackage{algpseudocode}
\algrenewcommand\algorithmicrequire{\textbf{Input:}}
\algrenewcommand\algorithmicensure{\textbf{Output:}}

\title{Large Causal Models from Large Language Models\thanks{Draft under revision.} }

\author{ Sridhar Mahadevan \\
	Adobe Research and University of Massachusetts, Amherst\\
	\texttt{smahadev@adobe.com, mahadeva@umass.edu}
}

\hypersetup{
pdftitle={A template for the arxiv style},
pdfsubject={q-bio.NC, q-bio.QM},
pdfauthor={David S.~Hippocampus, Elias D.~Striatum},
pdfkeywords={First keyword, Second keyword, More},
}

\begin{document}
\maketitle

\begin{abstract} 
We introduce a new paradigm for building large causal models (LCMs)  that exploits the enormous potential latent in today's large language models (LLMs). We describe our ongoing experiments with  an implemented system called DEMOCRITUS (Decentralized Extraction of Manifold Ontologies of Causal Relations Integrating  Topos Universal Slices) aimed at building, organizing, and visualizing LCMs that span disparate domains extracted from carefully targeted textual queries to LLMs. DEMOCRITUS is methodologically distinct from traditional narrow domain and hypothesis centered causal inference that builds causal models from experiments that produce numerical data.   A high-quality LLM (e.g.\ the $80$-billion parameter {\tt Qwen3-Next-80B-A3B-Instruct} \footnote{Specifically, we used a highly optimized Apple MLX version of {\tt Qwen3-Next-80B-A3B-Instruct} from Hugging Face in our experiments, downloadable from \url{https://huggingface.co/mlx-community/Qwen3-Next-80B-A3B-Instruct-6bit}}.) is used to propose topics, generate causal questions, and extract plausible causal statements from a diverse range of domains.  The technical challenge is then to take these isolated, fragmented, potentially ambiguous and possibly conflicting causal claims,  and weave them into a coherent whole,  converting them into relational causal triples and embedding them into a LCM. Addressing this technical challenge required inventing new categorical machine learning methods, which we can only briefly summarize in this paper, as it is focused more on the systems side of building DEMOCRITUS.  We describe the implementation pipeline for DEMOCRITUS comprising of six modules, examine its computational cost profile to determine where the current bottlenecks in scaling the system to larger models. We describe the results of using DEMOCRITUS over a wide range of domains, spanning archaeology, biology, climate change, economics, medicine and technology.  We discuss the limitations of the current DEMOCRITUS system, and outline directions for extending its capabilities. 
\end{abstract}

\keywords{Causal Discovery \and Large Language Models \and AI  \and Machine Learning}

\section{Introduction}

\begin{quote}
``I would rather discover one true cause than gain the kingdom of Persia" -- Democritus (460 -- 370 B.C.) 
\end{quote}

We introduce a new paradigm for building large causal models (LCMs)  that exploits the enormous potential latent in today's large language models (LLMs)  \citep{fm,deepseekai2025deepseekr1incentivizingreasoningcapability}. Much of the decades-long effort in causal discovery \citep{rubin-book,pearl-book,zanga2023surveycausaldiscoverytheory} has focused on constructing causal knowledge from carefully controlled highly specialized topically narrow studies in particular domains that typically yields numerical data. DEMOCRITUS is a methodologically distinct enterprise: build LCMs spanning potentially hundreds of distinct domains and ranging over millions of very specific causal claims, by carefully combining the vast knowledge latent in LLMs, with state-of-the-art categorical causal \citep{mahadevan2025intuitionisticjdocalculustoposcausal,Fritz_2020} and deep learning methods \citep{DBLP:conf/lics/FongST19,mahadevan2024gaiacategoricalfoundationsgenerative,gavranović2024positioncategoricaldeeplearning}. Our goal in this paper is to showcase the potential of DEMOCRITUS. To broaden the accessibility of this paper,  we omit a detailed discussion of much of the categorical machinery used in building DEMOCRITUS. We use an implemented version of the Geometric Transformer (GT), proposed originally in \citep{mahadevan2024gaiacategoricalfoundationsgenerative}, combined with a generalized backpropagation method that works by ``horn filling" gaps in simplicial sets \citep{DBLP:journals/entropy/Mahadevan23,may1992simplicial}, and several technical enhancements of these methods will be described in forthcoming papers. 

LLMs are increasingly capable of producing rich
causal narratives: they can enumerate subtopics, pose causal questions,
and articulate mechanistic explanations in domains ranging from
macroeconomics to neuroscience. However, using an LLM alone leaves us
with a \emph{laundry list} of disconnected fragments. Democritus aims
to turn these fragments into structured \emph{large causal models}--- which can be viewed categorically as slices of Topos Causal Models (TCMs) \citep{mahadevan2025intuitionisticjdocalculustoposcausal}
--- over which we can compute, visualize, and eventually reason. A {\em topos} is a type of category \citep{maclane:sheaves} that supports an internal intuitionistic logic. The use of such a logic was recently described in \citep{mahadevan2025intuitionisticjdocalculustoposcausal}. DEMOCRITUS, as described here, is purely a topos builder, but not (yet) a topos reasoner. One of the key strengths of TCMs is that they are robust to individual variability: they provide the theoretical basis for ``map-reduce" type decentralized approaches to causal discovery \citep{mahadevan2025decentralizedcausaldiscoveryusing}. This strength is fully utilized in DEMOCRITUS, which relies on combing through the vast nuggets of plausible causal statements generated by querying a modern state-of-the-art LLM to extract a coherent LCM. 

At a high level:
\begin{itemize}
\item A strong LLM (e.g., {\tt Qwen3-Next-80B-A3B-Instruct}) acts as a discovery engine
for domain topics, causal questions, and statements.
\item A Geometric Transformer (GT) layer runs over the resulting relational
graph and produces a manifold of node embeddings.
\item These manifolds are organized as slices of a larger topos, and
can be queried, visualized, and selectively refined.
\end{itemize}

A naive implementation expands every topic to a fixed depth, asks the
LLM for causal questions and statements at every node, and only then
builds the manifold. This works, but the LLM cost dominates and the
result is unfocused. In this paper, we first describe the basic
six-module pipeline, and then motivate and sketch an \emph{active}
version of Democritus in which feedback from Geometric Transformers
and downstream tasks guides where to explore next.

\begin{figure}[p]
    \centering
    \includegraphics[width=0.5\linewidth]{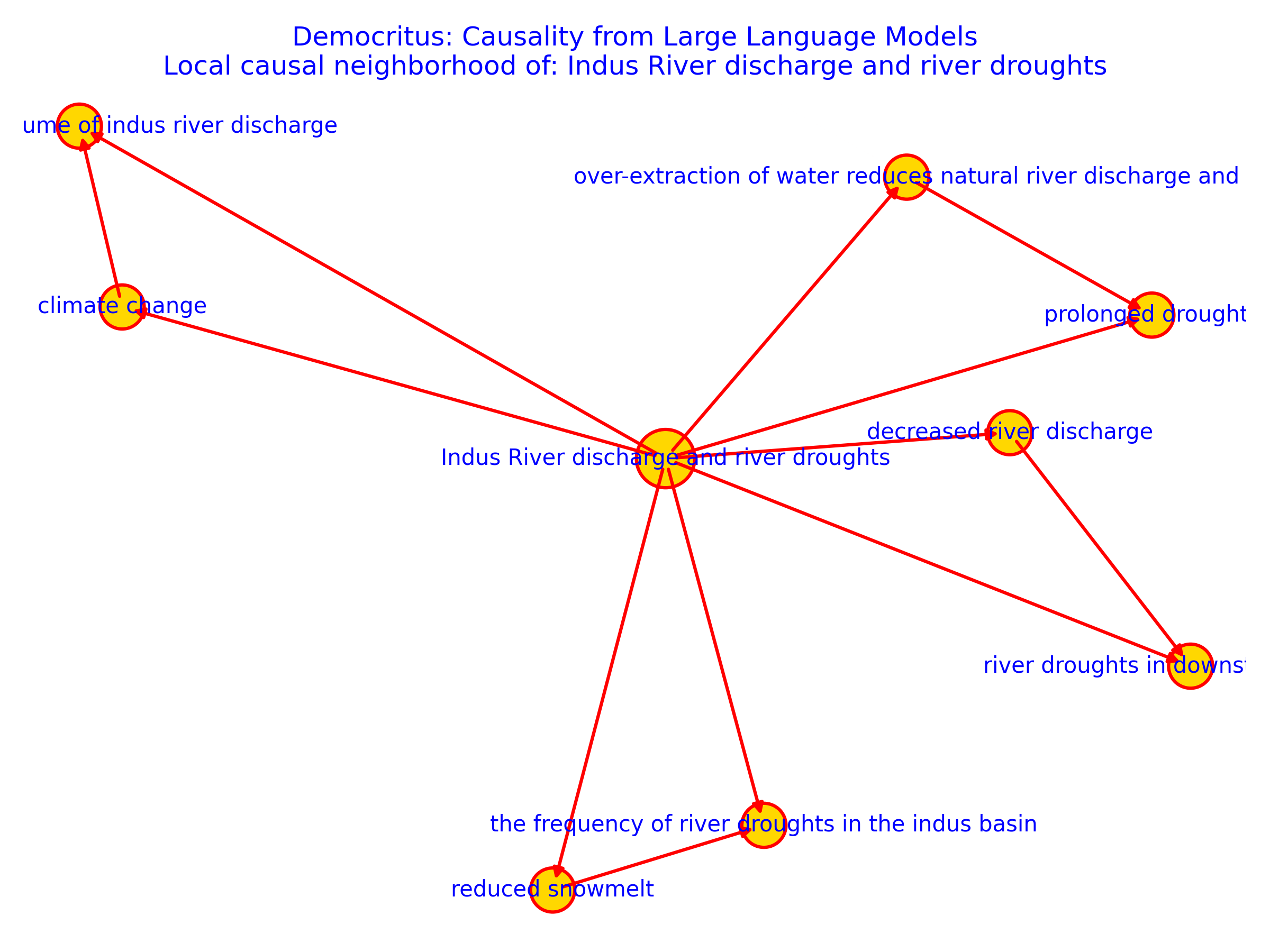}
    \caption{A local neighborhood causal model from a large LCM learned by DEMOCRITUS concerning the Indus river and droughts \citep{indus_valley_collapse}, which was used to explain the collapse of the Indus Valley Civilization 5000 years ago. Crucially, such a model cannot be learned by a single prompt to an LLM, but instead is woven together by an assembly of carefully curated prompts. }
    \label{fig:indus_river}
\end{figure}

\begin{figure}[p]
  \centering
  \includegraphics[width=\linewidth]{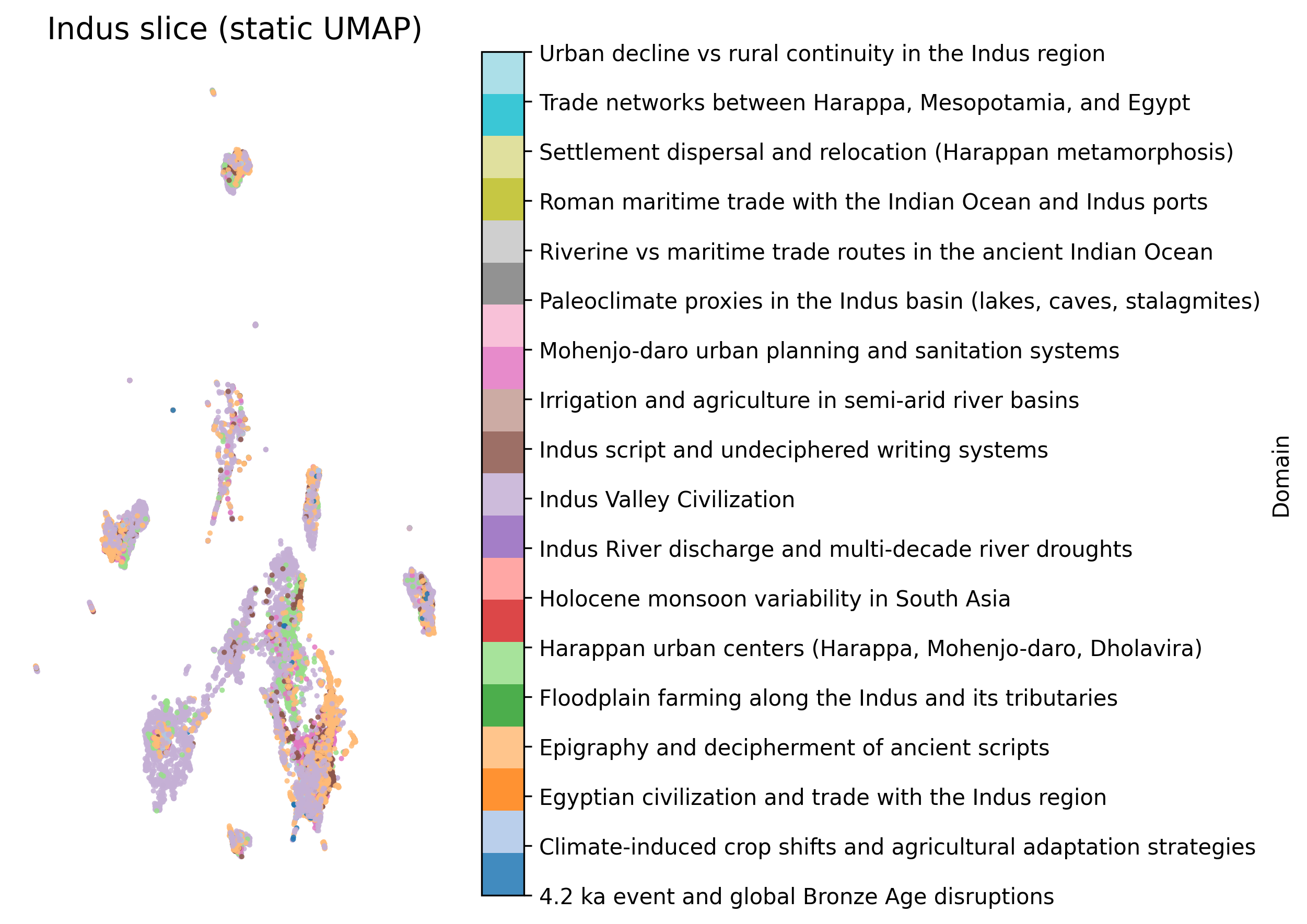}
  \caption{A global LCM underlying the collapse of the ancient Indus Valley civilization, shown as a 2D UMAP manifold of nodes colored
  by domain. Clusters correspond to Harappan urbanism and water
  systems, Indus script and epigraphy, climate and hydrology
  (Holocene monsoon variability, Indus river discharge and droughts),
  agriculture, and trade connections with Egypt and Rome, among
  others.}
  \label{fig:indus-manifold}
\end{figure}

As a showcase of the potential of DEMOCRITUS to elaborate on causal claims in the published literature, we apply it to a recent study \citep{indus_valley_collapse} that provides an explanation of the collapse of the Indus Valley civilization, which flourished on the banks of the River Indus about 5000 years ago. Such studies, typical in the literature across the humanities, sciences, and technology, require combining many fields of expertise, and serves as showcase that illustrates the strengths and limitations of DEMOCRITUS. 
Figure~\ref{fig:indus_river} illustrates a local causal model learned by DEMOCRITUS, whereas Figure~\ref{fig:indus-manifold} shows the entire global LCM, projected onto 2D using the UMAP data visualization method \citep{umap}. 

How did DEMOCRITUS end up with these local and global causal models?  As a way to ``prime the pump", we begin by giving DEMOCRITUS a set of plausible topics from which to generate causal claims including: 

\begin{itemize}
    \item Indus Valley Civilization / Harappan civilization
	\item 	Harappa, Mohenjo-daro, Dholavira, Ghaggar–Hakra (Saraswati)
	\item 	Holocene monsoon / paleoclimate / 4.2 ka event
	\item 	Indus river discharge, river droughts, streamflow anomalies
	\item 	Lake/playa shrinkage, stalagmites/stalagmites in Indian caves, lake sediments
	\item 	Wheat/barley/cotton agriculture, irrigation, floodplain farming
	\item 	Riverine trade networks (Harappa–Mesopotamia–Egypt), maritime vs river trade
	\item 	Settlement dispersal / relocation vs violent collapse
	\item 	4 long droughts, especially the ~164-year drought with ~13\% rainfall drop
\end{itemize}

DEMOCRITUS could prompt an LLM, such as {\tt Qwen3-Next-80B-A3B-Instruct}, with a prompt such as the following: 

\begin{tt}
    You are an expert in South Asian archaeology and paleoclimate. Given the topic ‘Indus Valley Civilization decline’, list 10 important subtopics that help explain its causes. Return only a numbered list, one per line.
\end{tt}

To seed the process of generating causal questions and statements, DEMOCRITUS can automatically generate causal questions, such as 

\begin{itemize}
    \item “What causes multi-decadal droughts in the Indus basin during the mid–late Holocene?”
	\item 	“How do changes in the tropical Pacific (El Niño/La Niña-like conditions) influence Indus monsoon rainfall?”
	\item 	“How does reduced monsoon rainfall propagate into river discharge, soil moisture, and crop viability?”
	\item 	“Why might repeated long droughts, rather than a single catastrophic event, drive gradual urban decline and dispersal?”
\end{itemize}

DEMOCRITUS then assembles the results of querying an LLM into ``causal triples", such as 

\begin{itemize}
    \item (tropical Pacific warming, reduces, monsoon rainfall)
    \item (reduced rainfall, decreases, Indus river discharge)
    \item (lower discharge, impedes, boat/barged river trade)
    \item (shrinkage of lakes/playas, reduces, local water storage)
    \item (multi-decade droughts, stress, agriculture and governance)
    \item (chronic water stress, leads\_to, relocation of Harappan settlements)
\end{itemize}

Our goal in doing this experiment is to explore how DEMOCRITUS can augment such published causal studies by filling in ``potential holes" in the findings, by combining the breadth of expertise in an LLM with a deep manifold extraction computation aided by categorical machine learning methods, such as the Geometric Transformer. For example, the list of possible ways in which such a study can be elaborated include: 

\begin{itemize} 
\item Graph connectivity: The paper focuses on climate → hydrology → trade/agriculture → settlement decline. Democritus might link to other well-known 4.2 ka event stories (Akkadian, Egyptian Old Kingdom, Caral), and tie paleoclimate proxies (caves, lakes, stalactites) to similar records elsewhere.

\item Alternative hypotheses / competing causes: DEMOCRITUS can enumerate other proposed drivers, including river avulsion and course changes, regional conflict, disease, trade shifts, internal social stratification.
These additional possibilities might let researchers see how these hypotheses sit in the manifold: near or far from the climate/hydrology cluster, which might reveal missing links.

\item Modern analogues: DEMOCRITUS can use the historical precedent of the Indus Valley collapse to explore potential modern impacts of climate change in the region. For example, it might suggest using the climate/hydrology structure in a modern South Asia slice (Ganges/Brahmaputra, current monsoon shifts), ask whether similar causal chains (multi-decade droughts, weak governance, river-dependent trade) appear today.

\item Active deepening: DEMOCRITUS can be used as a research tool by scientists to actively explore deeper into the LCM for a particular study, by first constructing a shallow Harappan topos slice first, then let the user “click” on a region (e.g. “multi-century river droughts”), and have Democritus selectively deepen around that region: more subtopics, more Q\&A, more triples.
\end{itemize}

\subsection{Limitations of \textsc{Democritus}}

A key limitation of the current version of \textsc{Democritus} is that the LCMs it proposes are not yet validated against numerical data or
controlled experiments. In this paper we focus on the \emph{hypothesis generation
and organization} problem: extracting and geometrically structuring candidate causal
edges from text. We do not attempt to estimate effect sizes or test these hypotheses
using observational or interventional datasets. In a future v2 system, we plan to
integrate \textsc{Democritus} with quantitative causal inference tools, so
that these edges can be assigned data-driven strengths and subjected to genuine
causal validation.

Even in the absence of numerical datasets, scientists routinely seek causal explanations
for events that cannot be experimentally replicated (e.g., the extinction of the
dinosaurs, the collapse of the Indus Valley civilization). In such domains, evidence
accumulates through the consilience of multiple partial traces: geological strata,
inscriptions, settlement patterns, and so on. \textsc{Democritus} v1 is designed
for exactly this setting: it aggregates and geometrically organizes textual causal
claims into a coherent hypothesis manifold, allowing researchers to see which mechanisms
recur across sources and where contradictions or gaps remain. In future work, we
aim to augment this textual aggregation with quantitative causal calculus when
suitable datasets or simulations are available.

We emphasize that \textsc{Democritus} does not attempt to replace formal causal
inference or identification strategies. The LCMs it constructs are best viewed
as structured \emph{hypothesis spaces} and \emph{narrative maps}: they make explicit what
the LLM \emph{implicitly} ``knows" about causal structure given its training data, but they
do not guarantee identifiability or causal correctness in the sense of
\citet{rubin-book,pearl-book}. Rather, they provide:
\begin{itemize}
\item a way to explore how different mechanisms and variables are linked in the
model's knowledge,
\item a source of candidate mechanisms and confounders for domain experts,
\item a geometric substrate on which more formal causal discovery methods could operate.
\end{itemize}

In this sense, \textsc{Democritus} is closer to a ``legal discovery'' or ``literature
review'' tool than to a parametric structural model. It organizes and visualizes
what an LLM already ``knows'' about a domain, and offers a starting point for human
experts and traditional causal tools to dig deeper.

\section{Related Work}
\label{sec:related}

DEMOCRITUS sits at the intersection of several research threads:
causal relation extraction from text, causal knowledge base
construction, large language models (LLMs) for causal discovery and
reasoning, and causality-aware NLP more broadly. We briefly review
each area and emphasize how DEMOCRITUS differs.

\subsection{Causal relation extraction from text}

There is a long line of work on identifying causal relations from
natural language, ranging from early pattern and cue-phrase methods
to modern neural models; see, for example, recent surveys on causal
relation extraction and event causality in text
(e.g.\ \citep{yang2022causalSurvey,he2022eventCausalitySurvey}).
Classical systems often rely on handcrafted patterns and lexical cues
such as ``because'' or ``leads to'', while more recent approaches use
supervised or semi-supervised neural architectures (including BERT
variants) trained to classify whether a pair of spans in a sentence
stands in a causal relation
\citep{radinsky2012learningCausality}.
These models typically operate at the level of individual sentences or
local contexts and predict pairwise labels.

In contrast, DEMOCRITUS assumes that an LLM can already express causal
relations in natural language, and uses the model as a generative
source of causal statements. The focus is not on marginally improving
pairwise classification accuracy, but on wiring thousands of such
statements into large, multi-hop causal graphs that can be explored
at different scales and embedded into manifolds.

\subsection{Causal knowledge bases and graphs from corpora}

Beyond pairwise extraction, several projects aim to construct causal
knowledge bases or causal graphs from large corpora
\citep{hassanzadeh2020causalKB}.
These systems mine causal tuples from text and aggregate them into
graph-structured resources (``causal KBs'', causal mind-maps, etc.).
Recent work has explored building such resources from specific domains
(e.g.\ maintenance logs, scholarly articles) and evaluating them on
downstream tasks.

DEMOCRITUS is similar in spirit, but differs in three ways. First, it
heavily exploits LLMs rather than classic pattern-based or supervised
models: Qwen3 is used to propose the variables, the causal questions,
and the causal narratives themselves. Second, the output is not only a
set of triples, but a navigable \emph{LCM}: nodes are
embedded via a Geometric Transformer and UMAP into 2D/3D spaces with
interpretable local neighborhoods. Third, DEMOCRITUS is explicitly
slice-based: causal graphs are organized as domain-specific slices
(econ, bio, Indus Valley, \ldots) inside a larger topos-style
architecture.

\subsection{Large language models for causal discovery and reasoning}

A rapidly growing literature investigates the use of LLMs for causal
discovery, causal reasoning, and explanation. Some works treat LLMs as
``meta-experts'' that can propose causal directions or graph structures
given textual descriptions of variables or data
\citep{shen2023llmCausalDiscovery}. Others use
multi-agent LLM setups to discuss and refine candidate DAGs
\citep{le2024multiagentCausalDiscovery}, or probe LLMs' internal
representations for causal notions such as interventions and
counterfactuals \citep{kosoy2023causalLLMReasoning}.

DEMOCRITUS occupies a complementary role. Rather than delegating causal
discovery to the LLM directly, we ask the model to articulate causal
\emph{propositions} in natural language (topics, questions,
statements), then treat these as noisy inputs to a separate causal
representation pipeline. The relational graphs and manifolds that
Democritus constructs can be used by classical causal discovery
algorithms or LLM-based reasoners, but the system itself is not a
causal learner in the sense of recovering ground-truth DAGs from
observational data.

\subsection{Causality and NLP more broadly}

More broadly, the intersection of causality and NLP has attracted
significant attention, including work on causal effects of text,
counterfactual data augmentation, and causal explanations for model
predictions; see, e.g., recent CausalNLP reading lists and surveys
\citep{jin2021causalnlpSurvey}. Democritus
contributes a different perspective: rather than using causal ideas to
\emph{improve} NLP models, it uses NLP (in the form of LLMs) to
construct explicit, structured causal artifacts that can be shared
across domains and inspected by humans and other systems.

\section{A Summary of How DEMOCRITUS Works}

We begin by summarizing how  DEMOCRITUS works and what it can produce, before getting into theoretical and implementation details. We show local causal models, which are simply neighborhoods near a causal variable, and manifold projections of LCMs learned in a number of domains. 

\subsection{A running example: influencer marketing}

\begin{figure}[t]
    \centering
    \includegraphics[width=0.6\linewidth]{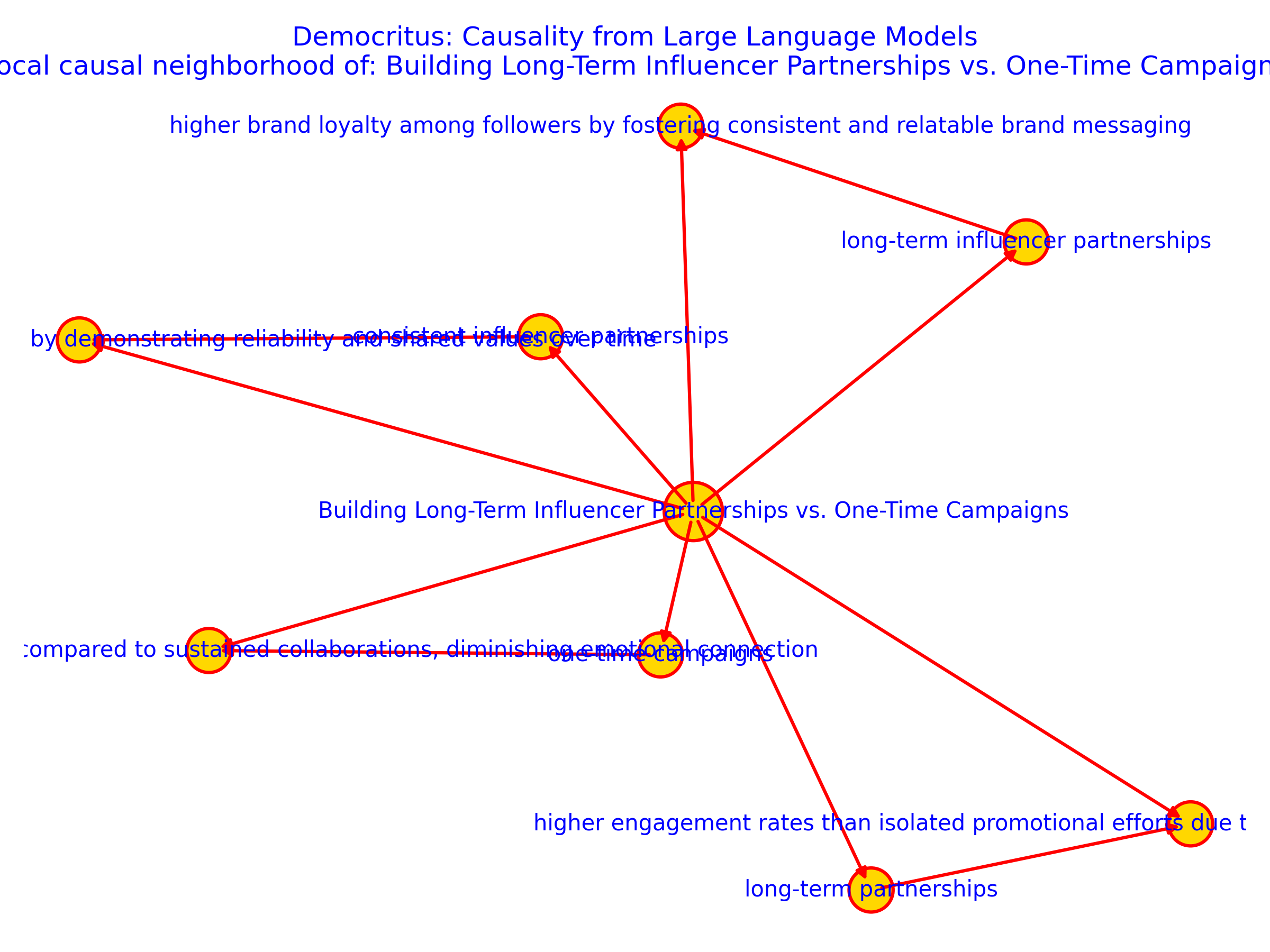}
     \caption{A neighborhood region of an LCM around the topic
  \emph{``Building Long-Term Influencer Partnerships vs.\ One-Time
  Campaigns''}. Nodes are variables/phrases; red arrows indicate causal
  directions; node size/colour reflects GT activations. This structure
  arises from DEMOCRITUS' multi-step pipeline (topic graph, causal
  statements, triple extraction, GT), not from a single prompt.}
    \label{fig:marketing-local}
\end{figure}

To illustrate what DEMOCRITUS builds, consider the neighborhood of an LCM
shown in Figure~\ref{fig:marketing-local}. The focus node is the topic
\emph{``Building Long-Term Influencer Partnerships vs.\ One-Time
Campaigns''}. Around this topic, DEMOCRITUS has constructed a small
directed causal graph that includes variables such as
\emph{``long-term influencer partnerships''}, \emph{``one-time
campaigns''}, \emph{``higher brand loyalty among followers by fostering
consistent and relatable brand messaging''}, \emph{``higher engagement
rates than isolated promotional efforts due to deeper audience
investment and repeated exposure''}, and
\emph{``the perceived authenticity of a brand compared to sustained
collaborations''}. Arrows indicate causal directions (e.g.\ long-term
partnerships \emph{lead to} higher engagement and brand loyalty),
and node size/colour reflects GT activations over this local graph.

Crucially, this structure is not the output of a single clever prompt
to an LLM. No one-shot query of the form
\emph{``Explain the causal effects of long-term influencer partnerships
versus one-time campaigns''} would return a navigable 2-hop graph with
typed edges, shared variables, and a geometric notion of salience. The
graph in Figure~\ref{fig:marketing-local} arises from the entire
DEMOCRITUS pipeline:

\begin{enumerate}
\item \textbf{Topic graph:} {\tt Qwen3-Next-80B-A3B-Instruct} is first asked, in Module~1, to
propose subtopics under marketing and influencer campaigns (e.g.\ long-
term partnerships, one-time campaigns, brand authenticity, audience
trust).
\item \textbf{Causal statements:} For each such topic, Modules~2--3
query {\tt Qwen3-Next-80B-A3B-Instruct} for short statements of the form \emph{``X causes Y''} or
\emph{``X leads to Y''} that describe causal relations in this niche.
\item \textbf{Triples and graph:} Module~4 extracts
(subject, relation, object) triples from these sentences and wires them
into a directed multi-relational graph whose nodes are phrases (e.g.\
``long-term influencer partnerships'', ``higher engagement rates'').
\item \textbf{Local neighborhood and GT:} Finally, we take the 2-hop
ego-graph around the focus node and run a Geometric Transformer layer
over it, yielding the local causal neighborhood and activations
visualised in Figure~\ref{fig:marketing-local}.
\end{enumerate}

In this sense, DEMOCRITUS is not simply a prompting technique; it is a
\emph{structure-building} system. The LLM supplies fragments of causal
knowledge, but the topic graph, triple extraction, and GT-based
manifold together turn those fragments into a reusable object: a local
causal model embedded in a larger economics topos slice. The rest of
the paper generalizes this story from influencer marketing to other
domains (macroeconomics, labor markets, biology, archaeology), and
shows how DEMOCRITUS can selectively deepen such models in response to
user tasks.

\subsection{Building LCMs using the Geometric Transformer and Topos Causal Models}
\label{sec:kan-do-gt-intro}

At the theoretical core of DEMOCRITUS lies two innovations: an implementation of the Geometric Transformer (GT) first theoretically conjectured in \citep{mahadevan2024gaiacategoricalfoundationsgenerative} whose implementation details will be described in a forthcoming paper \citep{mahadevan:gt-db}. Second, a framework for causal data integration was used that builds on our previous work on a decentralized map-reduce framework for learning Topos Causal Models \citep{mahadevan2025decentralizedcausaldiscoveryusing}.   We will describe each of these innovations in greater detail below, but briefly summarize the main ideas here and then show their results in a constructed LCM for several domains, ranging from biology to economics to Indus Valley archaeology. 

We ran DEMOCRITUS on 90,016 synthetic relational causal statements across 9 domains, from which a relational causal triples extracting module extracted 54,514 unique concepts and 57,390 typed relations.  The Geometric Transformer with Diagrammatic Backpropagation \citep{mahadevan:gt-db} then constructed a resulting multi-relational simplicial complex contains between 553 and 1,336 regime triangles per domain (approximately 9k 2-simplices in total). The 3D UMAP projection (Figure~\ref{fig:causal-map}) reveals coherent domain clusters with smooth transition zones exhibiting stratified domain sheets and cross-domain membranes that classical knowledge graph embeddings fail to capture.   Each sentence contains a linguistic relation (causes, influences, supports, prevents, is-a, part-of, etc.), and the model constructs a multi-relational simplicial complex. The  causal statements generated across 60 topical areas (economics, AI, public health, climate, demographics, etc.).  
Each statement is of the form
\[
    X \text{ increases } Y,\qquad
    X \text{ reduces } Y,\qquad
    X \text{ leads to } Y,
\]
and we extract causal triples to form a 1–simplicial causal graph.
Semantic domains supply 2–simplices, producing a causal site.

\subsection{LCMs: A Global View }

\begin{figure}[h]
\centering
\includegraphics[width=0.98\linewidth]{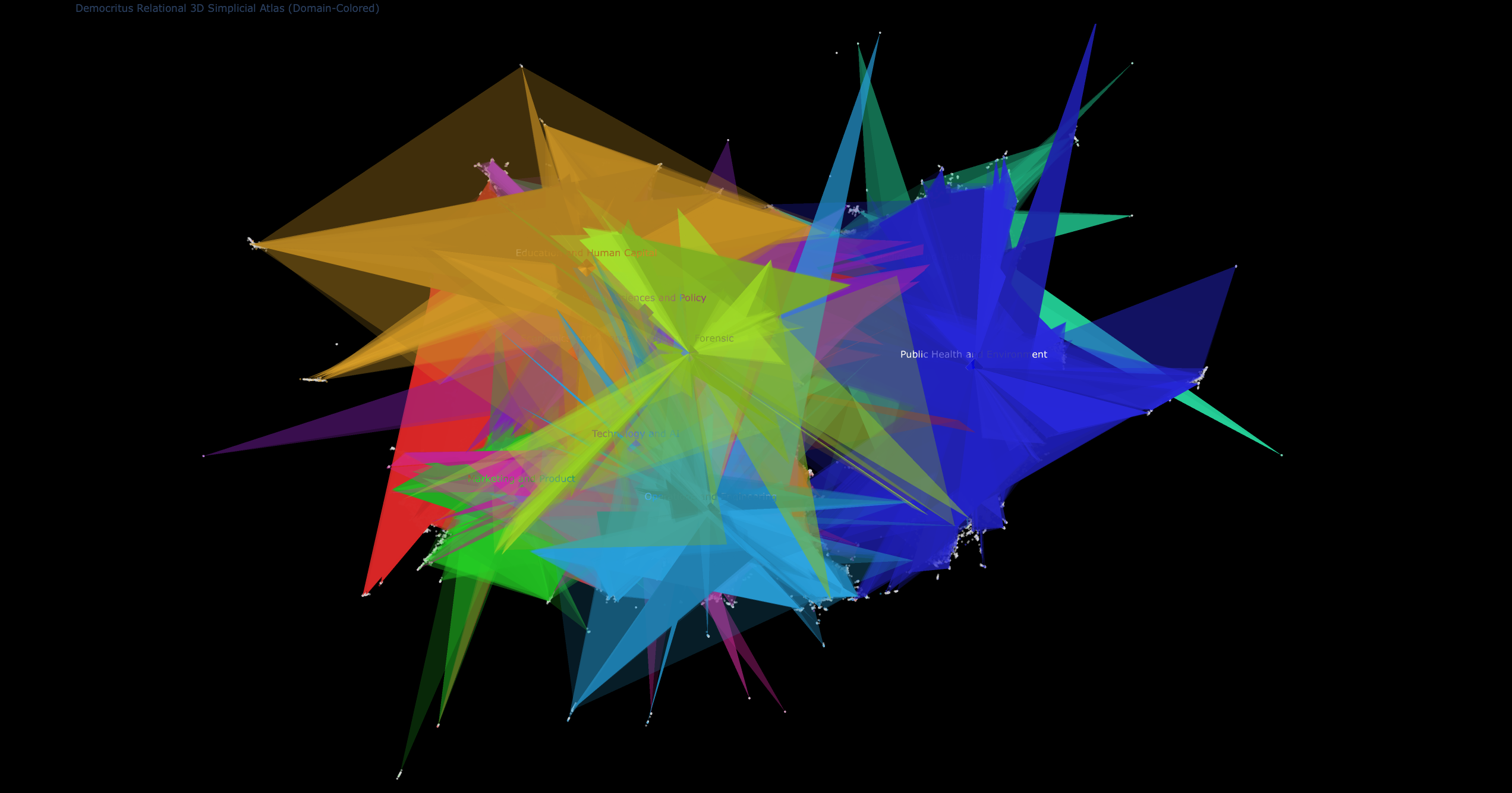}
\caption{
\textbf{Democritus LCM.}
3D UMAP projection of an LCM constructed from over $90,000$ causal textual statements sampled from GPT models in over $10$ domains.  The structure exhibits clear domain clustering and smooth causal transition regions.}
\label{fig:causal-map}
\end{figure}

Figure~\ref{fig:causal-map} shows the 3D UMAP embedding of the refined
diagrammatic causal geometry.   The manifold reveals several phenomena:

\begin{itemize}
\item \textbf{Macro-domain structure}: climate, inflation, vaccination,
      supply chains, mental health, and AI each form distinct regions.
\item \textbf{Causal gradients}: e.g.\ a clear transition from
      energy-related factors $\to$ electricity demand $\to$
      carbon emissions $\to$ climate impacts.
\item \textbf{Cross-domain interactions}: e.g.\ “generative AI” lies at
      a hub connecting education, productivity, misinformation, and
      creativity domains.
\end{itemize}

\subsection{Contrasting Domain and Relational LCMs}

\begin{figure}[h]
    \centering
    \includegraphics[width=0.5\linewidth]{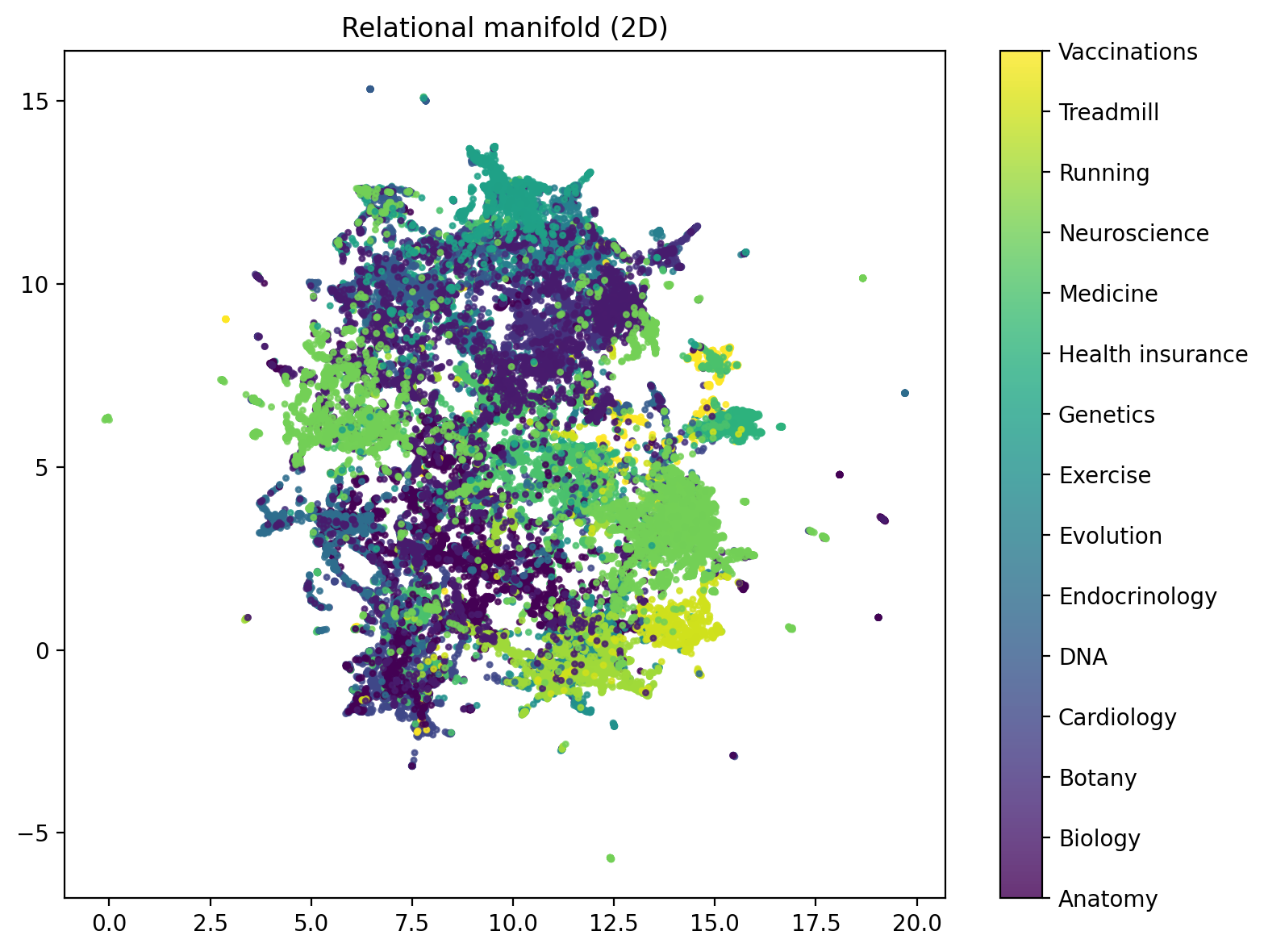}
    \caption{A 2D UMAP projection of an LCM for biology.}
    \label{fig:bio_global_manifold}
\end{figure}

\begin{figure}
    \centering
    \includegraphics[width=0.5\linewidth]{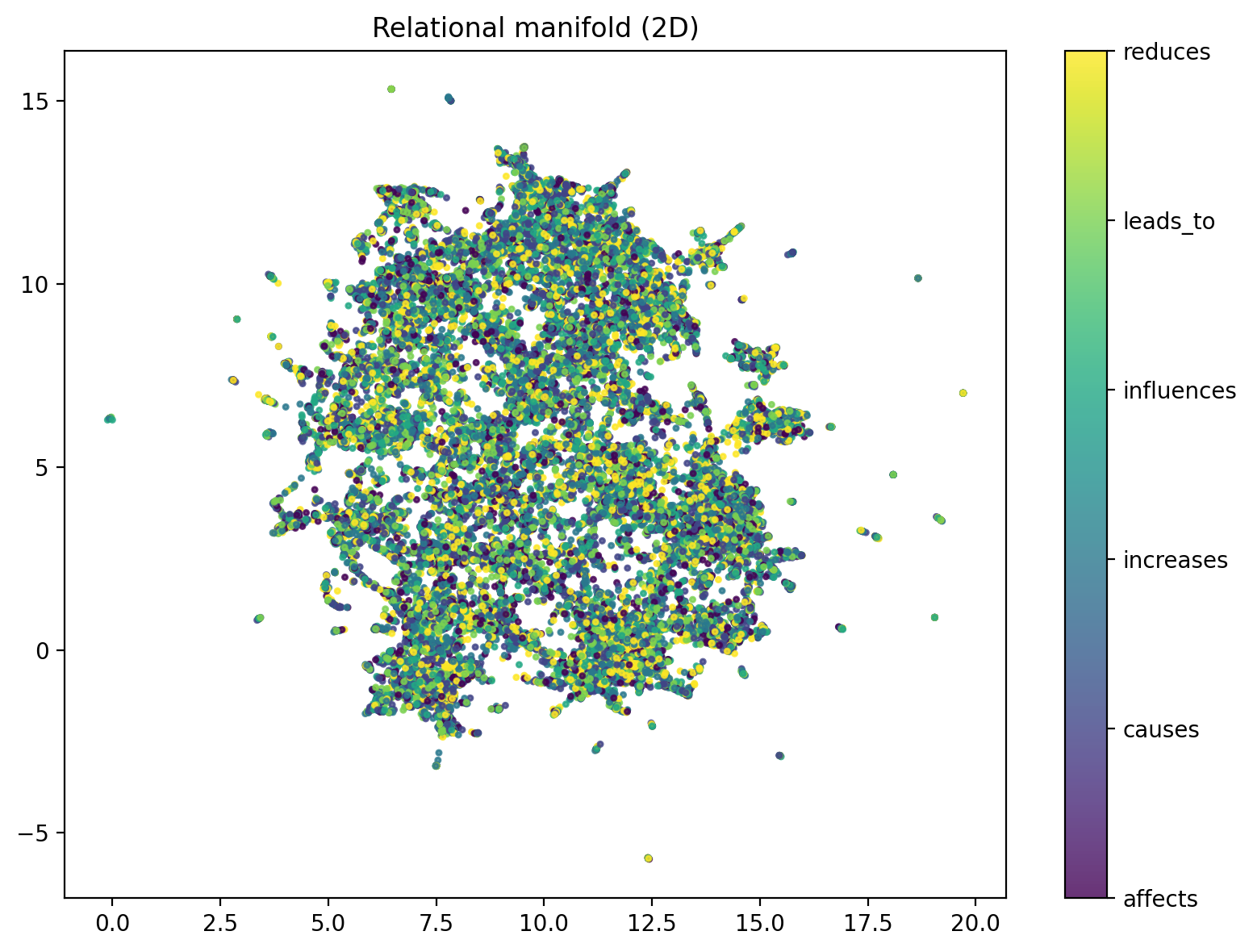}
    \caption{A relational LCM for a biology domain.}
    \label{fig:bio_relational_manifold}
\end{figure}

Figure~\ref{fig:bio_global_manifold} and Figure~\ref{fig:bio_relational_manifold} contrast two types of LCMs learned by DEMOCRITUS: one projects the causal data into the domain space, and the other projects the data onto the specific type of causal relationship. 

\subsection{LCMs: A Local Projection on Neighborhoods}
\begin{figure}[h]
\centering
\includegraphics[width=0.95\linewidth]{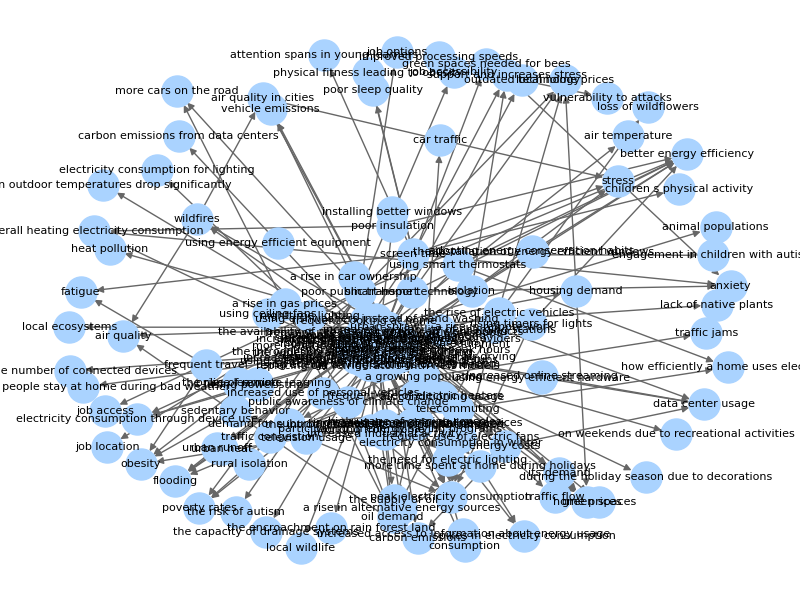}
\caption{
\textbf{Electricity demand causal neighborhood.}
Democritus recovers the natural causal cluster of heating/cooling,
EV charging, industrial activity, insulation, and demand-side behavior.}
\label{fig:neigh-electricity}
\end{figure}

\begin{figure}[h]
\centering
\includegraphics[width=0.95\linewidth]{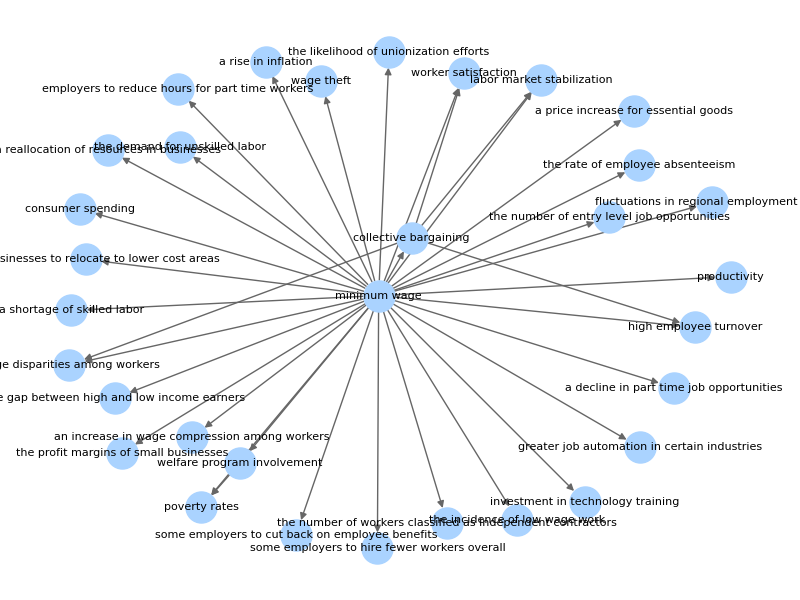}
\caption{
\textbf{Minimum wage causal neighborhood.}
A coherent causal cluster emerges: employment dynamics, consumer
spending, inflationary pressure, and labor productivity.}
\label{fig:neigh-minwage}
\end{figure}

Figures~\ref{fig:neigh-electricity}–\ref{fig:neigh-minwage} visualize
1–hop causal neighborhoods. These reveal that the learned LCM recovers
interpretable causal structure:

\begin{itemize}
\item Electricity demand is linked to heating, cooling, EV charging,
      industrial activity, and insulation quality.
\item Minimum wage connects to employment, consumer spending, inflation,
      and business closures.
\item Daily online users reveal strong ties to content quality, load
      balancing, advertising spend, and recommendation algorithms.
\end{itemize}

\subsection{Causal Degree Structure}
\begin{figure}[h]
\centering
\includegraphics[width=0.96\linewidth]{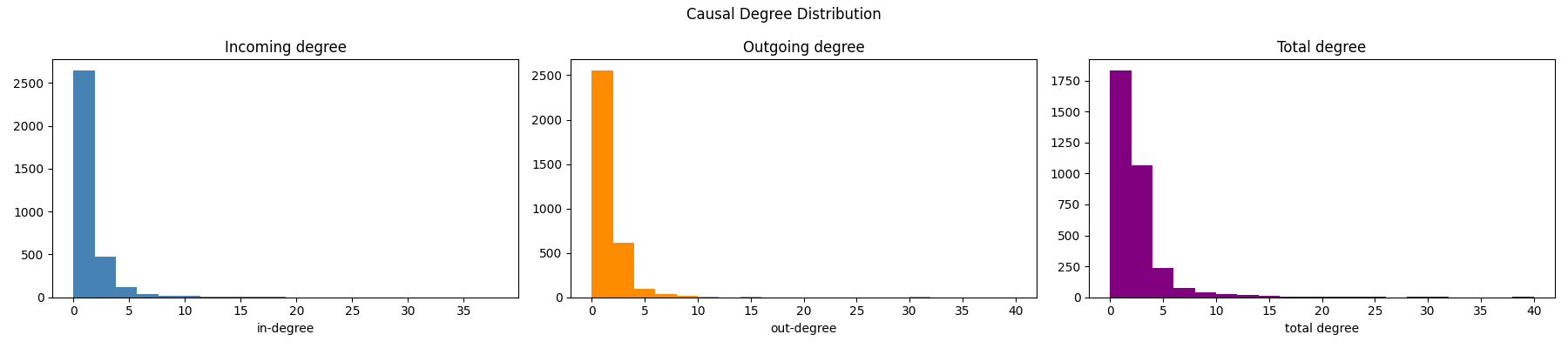}
\caption{
\textbf{Causal degree distribution.}
The causal graph exhibits a heavy-tailed distribution with recognizable
real-world “high-impact drivers’’ emerging as hubs.}
\label{fig:degree-dist}
\end{figure}

Figure~\ref{fig:degree-dist} shows the degree distribution of the causal
graph.  
The heavy-tailed structure suggests the existence of high-impact causal
variables such as stress, inflation, vaccination, generative AI, and
exercise — all of which emerge as central causal hubs in the manifold.

\subsection{Interpretation}
These experiments show that the learned LCMs produce a coherent system of causal
beliefs from raw textual statements.   The learned LCM demonstrates: 

\begin{itemize}
\item global causal coherence,
\item domain separation,
\item cross-domain bridges,
\item interpretable local neighborhoods,
\item and causal hubs that resemble domain expertise.
\end{itemize}

\section{System overview}

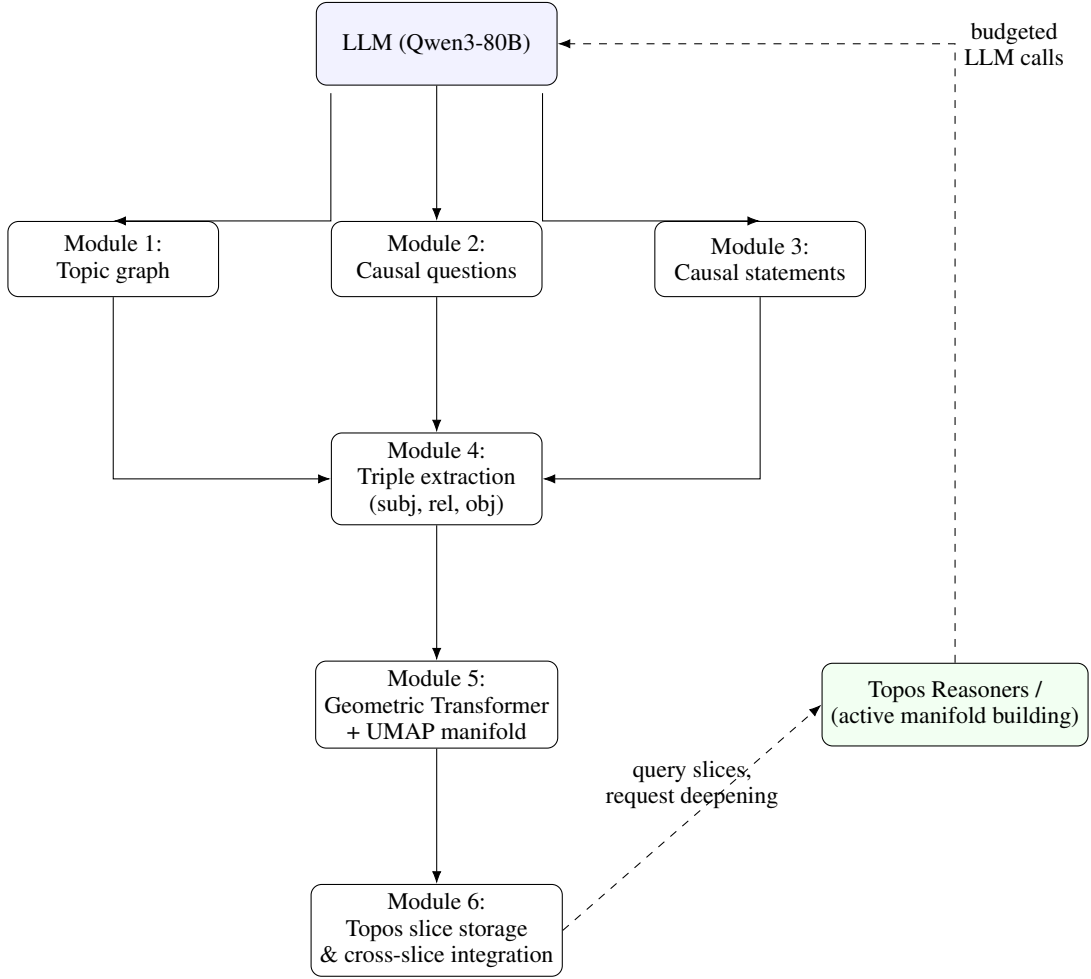
\begin{figure}[t]
  \centering
  \begin{tikzpicture}[
      node distance=1.8cm and 1.6cm,
      >=Latex,
      every node/.style={font=\small},
      box/.style={rectangle, draw, rounded corners, align=center, minimum width=2.8cm, minimum height=1.0cm},
      bigbox/.style={rectangle, draw, rounded corners, align=center, minimum width=3.2cm, minimum height=1.1cm},
    ]

    \node[bigbox, fill=blue!5] (llm) {LLM (Qwen3-80B)};

    \node[box, below=of llm, xshift=-4.3cm] (m1) {Module 1:\\ Topic graph};
    \node[box, below=of llm]                (m2) {Module 2:\\ Causal questions};
    \node[box, below=of llm, xshift=4.3cm]  (m3) {Module 3:\\ Causal statements};

    \draw[->] (llm.south west)++(0.2,-0.1) |- (m1.north);
    \draw[->] (llm)                          -- (m2.north);
    \draw[->] (llm.south east)++(-0.2,-0.1) |- (m3.north);

    \node[box, below=1.8cm of m2] (m4) {Module 4:\\ Triple extraction\\ (subj, rel, obj)};

    \draw[->] (m1.south) |- (m4.west);
    \draw[->] (m2.south) -- (m4.north);
    \draw[->] (m3.south) |- (m4.east);

    \node[box, below=1.8cm of m4] (m5) {Module 5:\\ Geometric Transformer\\ + UMAP manifold};

    \draw[->] (m4.south) -- (m5.north);

    \node[box, below=1.8cm of m5] (m6) {Module 6:\\ Topos slice storage\\ \& cross-slice integration};

    \draw[->] (m5.south) -- (m6.north);

    \node[bigbox, right=3.5cm of m5, fill=green!5] (ctrl) {Topos Reasoners / \\ (active manifold building)};

    \draw[->, dashed] (m6.east) -- node[above,align=center]{query slices,\\ request deepening} (ctrl.west);
    \draw[->, dashed] (ctrl.north) |- node[right,align=center]{budgeted\\ LLM calls} (llm.east);

  \end{tikzpicture}
  \caption{High-level DEMOCRITUS pipeline. Modules~1--3 use the LLM to
  propose topics, causal questions, and statements. Module~4 extracts
  triples and builds a relational graph. Module~5 applies a Geometric
  Transformer and UMAP to obtain a LCM. Module~6 stores
  domain-specific slices and supports cross-slice integration. A
  topos reasoner (e.g., Judo calculus \citep{mahadevan2025intuitionisticjdocalculustoposcausal}) can be used to actively query slices and request
  further LLM calls in selected regions.}
  \label{fig:democritus-pipeline}
\end{figure}

Figure~\ref{fig:democritus-pipeline} shows the overall
Democritus pipeline for a single domain slice (e.g.\ economics or
biology). The modules are:

\begin{enumerate}
\item \textbf{Topic graph (Module 1).} Build a topic hierarchy
via LLM-driven breadth-first expansion.
\item \textbf{Causal questions (Module 2).} For each topic, generate
causal questions.
\item \textbf{Causal statements (Module 3).} For each topic, generate
causal statements or explanations.
\item \textbf{Relational triples (Module 4).} Extract subject--relation--object
triples from questions and statements.
\item \textbf{Relational manifold (Module 5).} Embed the relational graph
with a Geometric Transformer, and compute a low-dimensional manifold
(2D/3D) for visualization.
\item \textbf{Topos slice and unification (Module 6).} Store the resulting
slice, optionally unify multiple slices, and make them available to
downstream intuitionistic causal reasoners using Judo calculus  \citep{mahadevan2025intuitionisticjdocalculustoposcausal}. 
\end{enumerate}

In the current implementation, each LCM is built independently for a
given domain (e.g.\ economics, biology), and the Geometric Transformer operates at the level
of each slice. Figure~\ref{fig:democritus-pipeline} gives a high-level overview of the DEMOCRITUS pipeline. We briefly describe each module, illustrating with the economics (econ)
and biology (bio) slices constructed in our current implementation.

\subsection{Module 1: Topic graph}

Given a set of root topics (e.g.\ \texttt{Macroeconomics,
Microeconomics, Game Theory, Finance, Trade, Marketing, Stock Market,
Investing, Cryptocurrency, Bonds, Monetary Policy, Banking, Fiscal
Policy, Inflation, Unemployment}), we perform a breadth-first search
(BFS) expansion using Qwen3-Next-80B-A3B-Instruct-6bit. At each node
the LLM is prompted to list a small number of subtopics; these become
children in the topic graph.

A typical run with depth limit 5 and maximum 7000 topics in the domain of economics yields
a rich hierarchy including subtrees for monetary policy, game theory,
cryptocurrency/DeFi, bond markets, agricultural subsidies, and so on.
An analogous bio run with roots such as \texttt{Neuroscience,
Genetics, Evolution, Botany, Cardiology} yields a similarly rich
hierarchy.

The output of this module is:
\begin{itemize}
\item a JSONL topic graph (\texttt{topic\_graph\_econ.jsonl},
\texttt{topic\_graph\_bio.jsonl}),
\item a tabular topic list with depths (\texttt{topic\_list\_econ.txt},
\texttt{topic\_list\_bio.txt}).
\end{itemize}

\subsection{Modules 2 and 3: Causal questions and statements}

For each topic in the graph, we prompt Qwen3 to generate:
\begin{itemize}
\item causal questions (Module~2), e.g.\ \emph{``What causes a rise in
the price of gold?''}, \emph{``What leads to a leftward shift in the
demand curve?''}
\item causal statements or explanations (Module~3), e.g.\ 
\emph{``Increased demand for gold as a safe-haven asset during economic
uncertainty causes its price to rise.''}
\end{itemize}

These are stored as JSONL files
(\texttt{causal\_questions.jsonl}, \texttt{causal\_statements.jsonl})
with fields for topic, path, question, and an array of statements. For
example, an econ entry might read:
\begin{quote}
\small
\texttt{\{"topic": "Macroeconomics", "path": ["Macroeconomics"],}\\
\texttt{"question": "What causes a rise in the price of gold?",}\\
\texttt{"statements": ["Increased demand for gold as a safe-haven}\\
\texttt{asset during economic uncertainty causes its price to rise.",}\\
\texttt{"A decline in the value of the U.S. dollar leads to higher}\\
\texttt{gold prices due to its inverse correlation.", ...]\}}
\end{quote}

\subsection{Module 4: Relational triples}

From questions and statements we extract relational triples of the form
\((\text{subj}, \text{rel}, \text{obj})\), e.g.
\begin{center}
\(\text{``demand for gold''},\ \text{causes},\ \text{``price of gold rises''}\).
\end{center}
These triples define a directed multi-relational graph over phrases or
variables with relation types such as \texttt{causes},
\texttt{increases}, \texttt{influences}, \texttt{leads\_to},
\texttt{reduces}. In practice we obtain hundreds to thousands of
triples even for small runs (e.g.\ depth 2, 100 topics), and many more
for larger runs.

\subsection{Module 5: Refining the LCM via the Geometric Transformer}

We then embed the relational graph using a Geometric Transformer \citep{mahadevan2024gaiacategoricalfoundationsgenerative}. Nodes carry initial text-based embeddings from
sentence encoders or LLM-derived vectors; GT passes messages along
edges (1-simplices) and optionally across higher-order motifs (e.g.
triangles as 2-simplices), and produces refined node embeddings. A
low-dimensional manifold is obtained via UMAP. Even small runs show dense, semantically coherent
clusters; larger runs (e.g.\ 7000 topics at depth 5) yield richer,
thicker models.

\subsection{Module 6: Topos slices and unification}

Each domain (econ, bio, etc.) yields a slice: a relational graph with
structured embeddings, visualizations, and metadata. These slices are stored as
persistent objects,  and can be unified or queried jointly.
Controllers built around topos reasoners, such as Judo Calculus \citep{mahadevan2025intuitionisticjdocalculustoposcausal},  can treat these slices as structured
external memory, asking to deepen specific regions or relate patterns
across domains.

\begin{figure}
    \centering
    \includegraphics[width=0.5\linewidth]{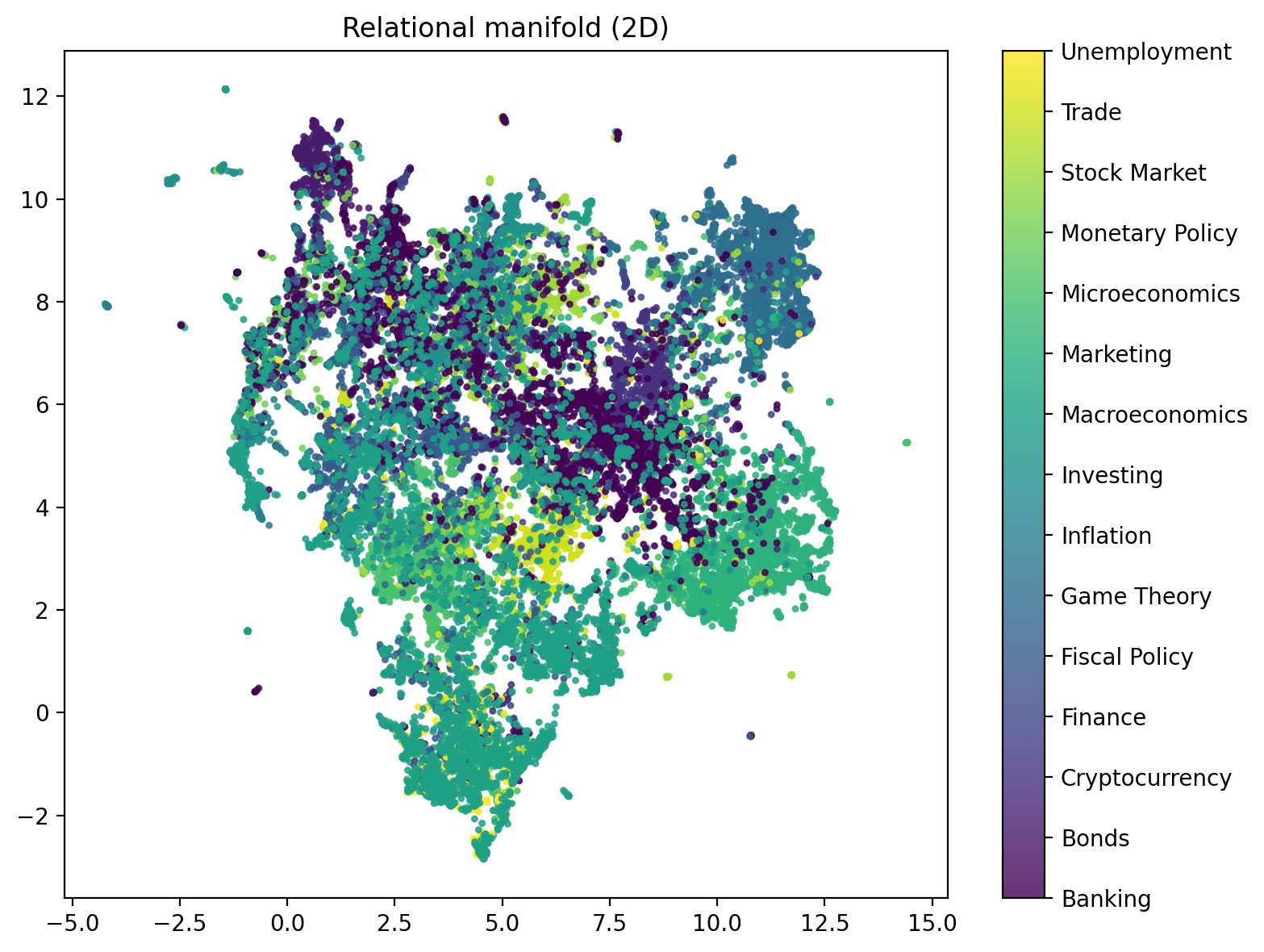}
    \caption{The global LCM for an economics domain over  $ \sim 35,000$ variables.}
    \label{fig:econ_global}
\end{figure}

Figure~\ref{fig:econ_global} shows the result of running all the six modules on an economics domain using the {\tt Qwen3-Next-80B-A3B-Instruct} LLM, where the resulting manifold is over almost $\sim 35,000$ variables. 

\section{A Walk Through of the DEMOCRITUS Pipeline}
\label{sec:walkthrough}

We highlight a single path through this pipeline, starting from the
topic \texttt{Macroeconomics} and ending in a local causal neighborhood
around gold prices in the global manifold.

\subsection*{1. Topic expansion (Module 1)}

Given an initial list of root topics, DEMOCRITUS asks {\tt Qwen3-Next-80B-A3B-Instruct} to
propose subtopics for each parent. For example, for the root topic
\texttt{Macroeconomics} we issue a prompt of the form:
\begin{quote}
\small
\textbf{Prompt (topic expansion).}\\[2pt]
\texttt{You are an expert economics textbook author.}\\
\texttt{Given the topic "Macroeconomics", list 10 important subtopics.}\\
\texttt{Return only a numbered list of subtopics, one per line.}
\end{quote}
{\tt Qwen3-Next-80B-A3B-Instruct} responds with subtopics such as:
\begin{quote}
\small
\texttt{1. Gross Domestic Product (GDP) and its measurement}\\
\texttt{2. Inflation and price indices}\\
\texttt{3. Unemployment types and rates}\\
\texttt{4. Fiscal policy and government spending}\\
\texttt{5. Monetary policy and central banking}\\
\texttt{6. Aggregate demand and aggregate supply}\\
\texttt{7. Business cycles and economic fluctuations}\\
\texttt{8. International trade and balance of payments}\\
\texttt{9. Exchange rates and foreign exchange markets}\\
\texttt{10. Long-run economic growth and productivity}
\end{quote}
Each line becomes a child node in the topic graph, with a depth counter
(0 for roots, 1 for their children, etc.). Similar prompts expand
subtopics under \texttt{Microeconomics}, \texttt{Game Theory},
\texttt{Finance}, and so on, yielding a directed acyclic graph of
several thousand topics at depths up to~5.

\subsection*{2. Causal questions and statements (Modules 2 and 3)}

For each topic in the graph, DEMOCRITUS next asks {\tt Qwen3-Next-80B-A3B-Instruct} to generate
causal questions and statements. For example, for the topic
\texttt{Macroeconomics} we might ask:
\begin{quote}
\small
\textbf{Prompt (causal questions).}\\[2pt]
\texttt{You are an expert in macroeconomics.}\\
\texttt{Topic: "Macroeconomics".}\\
\texttt{Write 3 causal questions a student might ask about this topic.}\\
\texttt{Each question should start with "What causes" or "What leads to".}\\[4pt]
\textbf{Prompt (causal statements).}\\[2pt]
\texttt{For the topic "Macroeconomics", write 3 short statements}\\
\texttt{of the form "X causes Y" or "X leads to Y" that describe}\\
\texttt{causal relationships in this topic.}
\end{quote}

The model might respond with a question such as:
\begin{quote}
\small
\emph{``What causes a rise in the price of gold?''}
\end{quote}
and statements such as:
\begin{quote}
\small
\emph{``Increased demand for gold as a safe-haven asset during economic}\\
\emph{uncertainty causes its price to rise.''}\\
\emph{``A decline in the value of the U.S. dollar leads to higher gold}\\
\emph{prices due to its inverse correlation.''}\\
\emph{``Geopolitical instability influences investor behavior, resulting}\\
\emph{in greater gold accumulation and higher prices.''}
\end{quote}
Similar prompts are issued for hundreds or thousands of topics (e.g.\
\texttt{Inflation}, \texttt{Unemployment}, \texttt{Fiscal Policy},
\texttt{Cryptocurrency}, \texttt{Systemic risk identification and
monitoring frameworks}, etc.), producing a large collection of causal
sentences.

\subsection*{3. Triples and causal graphs (Module 4)}

From the questions and statements, DEMOCRITUS extracts relational
triples of the form (subject, relation, object). For the gold example
above, this yields entries such as:
\begin{align*}
  &(\text{``demand for gold as a safe-haven asset''},\ \texttt{causes},\ \text{``price of gold rises''}),\\
  &(\text{``decline in the value of the U.S. dollar''},\ \texttt{leads\_to},\ \text{``higher gold prices''}),\\
  &(\text{``geopolitical instability''},\ \texttt{influences},\ \text{``gold accumulation and prices''}).
\end{align*}
Collecting triples across all topics yields a directed multi-relational
graph whose nodes are phrases/variables and whose edges carry relation
types such as \texttt{causes}, \texttt{increases}, \texttt{influences},
\texttt{leads\_to}, \texttt{reduces}. Even a small run at depth~2 with
100 topics produces several hundred such nodes and edges; a larger run
at depth~5 with 7000 topics yields tens of thousands.

\subsection*{4. Manifold embedding and local views (Module 5)}

To organize this graph, DEMOCRITUS applies a Geometric Transformer
layer over the triples. Nodes are initialized with text-based
embeddings; GT passes messages along edges and across higher-order
motifs (e.g.\ triangles of variables), producing refined node
embeddings. Applying UMAP to these embeddings yields a 2D/3D  visualization.

\section{Causal graph construction and manifold refinement}
\label{sec:manifold-refinement}

In this section we briefly describe how DEMOCRITUS turns LLM-derived
triples into a refined LCM. Our goal is to explain the
process we actually use in the system, without developing the full
theoretical machinery, which can be found in our earlier publications~\citep{mahadevan2024gaiacategoricalfoundationsgenerative}.

\subsection{Relational causal graph construction}
\label{subsec:relational-graph}

After Modules~1--3 have generated causal questions and statements for a
given slice (e.g., economics, biology, Indus Valley), Module~4 extracts
typed triples of the form $(\text{head}, \text{relation}, \text{tail})$
along with a domain label, following the standard OpenIE-style
parsing. We then build a variable-level relational graph:

\begin{itemize}
\item Each distinct head or tail string becomes a variable node $v \in V$
      (e.g.\ ``chronic stress'', ``cardiovascular risk'').
\item For each triple $(h,r,t)$ we add a directed edge $h \to t$ of
      relation type $r$ and domain label $d$ (e.g.\ biology, economics).
\item This yields a multi-relational directed graph
      $G = (V, E, \text{rel}, \text{dom})$ which aggregates causal
      statements across all topics in the slice.
\end{itemize}

Edge multiplicities and domain labels encode how often and in which
subdomains a given mechanism is asserted. The result is a large,
sparse, heavy-tailed causal graph (Section~\ref{subsec:bio-spectral-structure})
with a few high-degree variables (hubs) and many low-degree fringe
nodes.

\subsection{Higher-order message passing}
\label{subsec:message-passing}

To refine and denoise the raw relational graph, we apply a small
higher-order message-passing network before running UMAP. In our
current implementation, DEMOCRITUS uses a lightweight two-layer
architecture with both edge-level and triangle-level aggregation:

\begin{itemize}
\item Each variable node $v$ starts with an initial embedding
      $\mathbf{h}_v^{(0)}$ obtained from a sentence encoder applied to
      the variable string (we use a standard transformer-based encoder).
\item At each message-passing layer $\ell{+}1$, we update
      $\mathbf{h}_v^{(\ell+1)}$ by aggregating messages from its
      neighbours $u$ along edges $u \to v$, weighted by learned
      embeddings of the relation type and domain label. This is a
      standard edge-based message-passing step.
\item In addition, we detect simple higher-order motifs (triangles)
      where $(u \to v)$ and $(v \to w)$ share the same domain; these
      define 2-simplices $(u,v,w)$ in the graph. We aggregate messages
      over such triangles and add them as an extra term to the update,
      so variables that participate in many coherent motifs are nudged
      towards each other in the embedding space.
\end{itemize}

This results in a refined embedding $\mathbf{h}_v$ for each variable
that smooths over local inconsistencies and takes higher-order structure
into account (e.g.\ chains and triangles of causal influence), without
requiring any labeled supervision. Conceptually, this is a simple
instantiation of higher-order geometric message passing over a causal
graph; more sophisticated versions are possible, but we find this
lightweight architecture sufficient for the slices considered here.

\subsection{Manifold embedding and visualization}
\label{subsec:umap-embedding}

Finally, we run UMAP on the refined embeddings $\{\mathbf{h}_v\}_{v \in V}$
to obtain two- and three-dimensional coordinates for each variable.
The resulting low-dimensional manifold exhibits coherent clusters
corresponding to subdomains and mechanisms (e.g.\ stress and
cardiometabolic risk, Indus monsoon and river discharge).  In
Section~\ref{sec:manifold-robustness} we examine the degree
distributions, Laplacian spectra, and stability of these graphs and
manifolds, and show that the resulting structures are sparse,
heavy-tailed, and robust to modest noise and repeated runs.

\section{Comparative Experiments}

We report on some preliminary baseline experiments that show the effect of using the Geometric Transformer in constructing LCMs, as opposed to a naive use of UMAP to directly construct embeddings. 

\subsection{Baseline Experiment 1: Modules 1 through 4 only}

\begin{figure}
    \centering
    \includegraphics[width=0.5\linewidth]{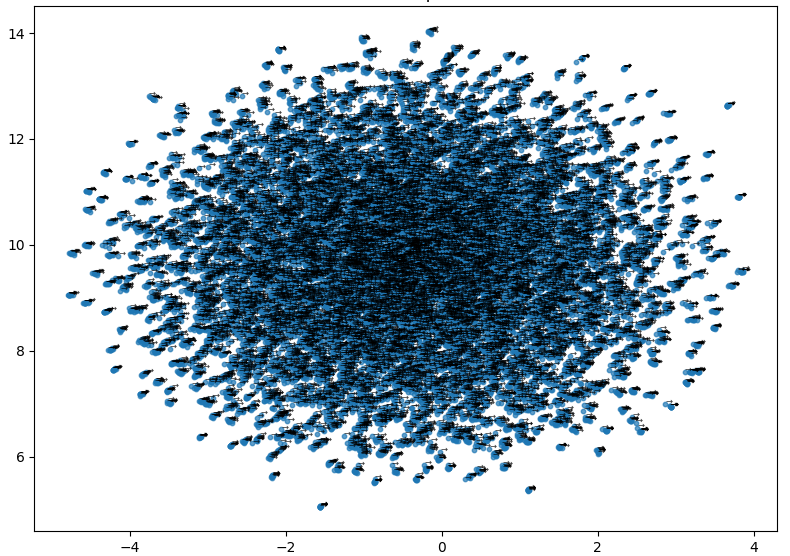}
    \caption{The unstructured LCM that results from running UMAP on causal triples over $\sim 35,000$ variables, without using the Geometric Transformer.}
    \label{fig:umap_only_no_gt}
\end{figure}

Suppose we just use UMAP to construct manifold embeddings of the relational triples extracted from Modules 1 through 4.  Does that reveal the same structure? Figure~\ref{fig:umap_only_no_gt} shows that without the Geometric Transformer, the UMAP-generated embeddings are a like a ``giant hairball" with no discernible structure. This experiment used the economics domain, with almost $35,000$ causal variables extracted from a run on {\tt Qwen3-80B}. Clearly, comparing the plot produced earlier in Figure~\ref{fig:econ_global}, we can see that UMAP alone is unable to discover the deep structure of the global economics LCM, even given the same exact relational triples embedded in a high-dimensional space using a Sentence Transformer encoder like BERT. 

\subsection{Baseline Experiment 2: Adding the Geometric Transformer only}

\begin{figure}
    \centering
    \includegraphics[width=0.5\linewidth]{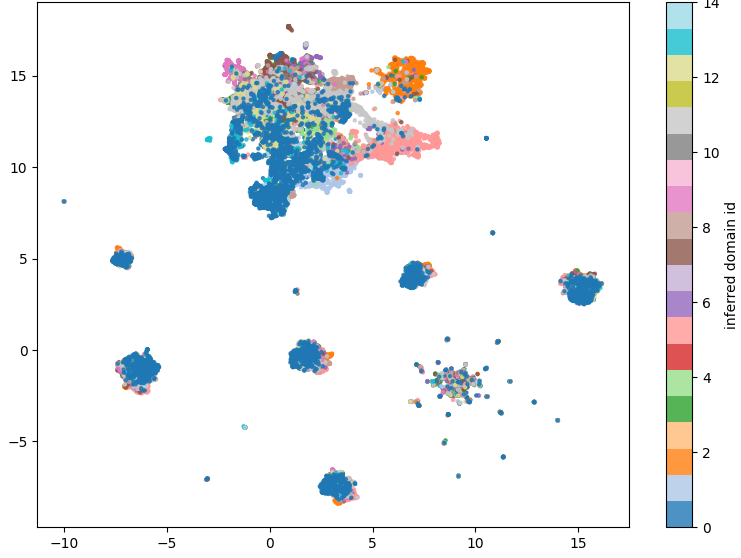}
    \caption{The LCM structure discovered by Modules 1 through 5, including the Geometric Transformer refinement.}
    \label{fig:gt_only}
\end{figure}

Now,  we extend the previous baseline experiment of running Modules 1 through 4 only and producing the UMAP embedding shown in Figure~\ref{fig:umap_only_no_gt} by adding the Geometric Transformer in Module 5. Figure~\ref{fig:gt_only} shows the results, again over the economics domain of $\sim 35,000$ variables. Now, we see the effect of the Geometric Transformer, where the structural aspects of the economics LCM snap into sharp relief. There are clearly discernible structures across the various subfields of economics. 

\subsection{Baseline Experiment 3: Adding the Causal Refinement Step}

\begin{figure}
    \centering
    \includegraphics[width=0.5\linewidth]{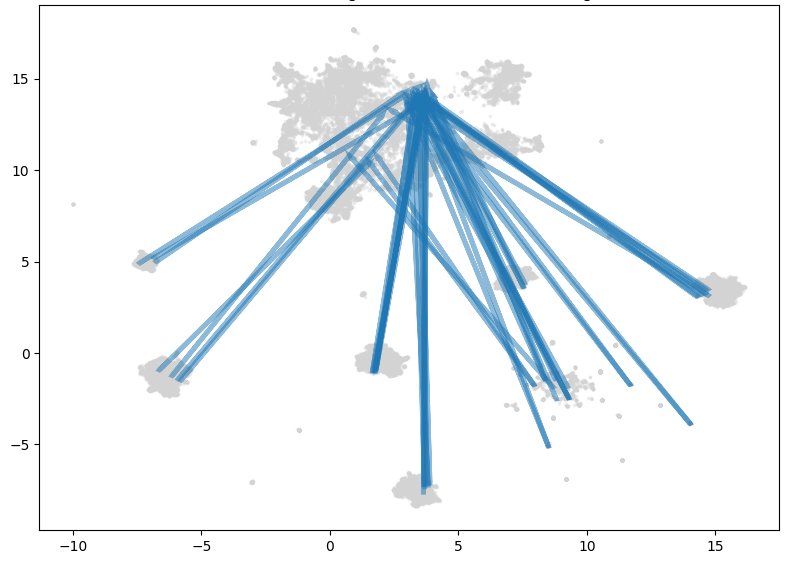}
    \caption{This figure shows the effect of the causal refinement layer, which brings in causal inference to refine the $\sim 35,000$ economics LCM.}
    \label{fig:econ_gt_kan_do_expt}
\end{figure}

To complete the baseline experiments, we now show the effect of the causal refinement step, where we now begin to introduce actual causal experimental on the manifold structure shown in Figure~\ref{fig:gt_only}. The result is shown in Figure~\ref{fig:econ_gt_kan_do_expt}. Now we begin to see edges form between major hubs in the economics manifold. These edges correspond to real causal effects, such as: 

\begin{verbatim}
government spending -> an increase in aggregate demand by injecting more money into the economy
inflation -> the purchasing power of consumers by raising the prices of goods and services
unemployment -> overall economic growth by lowering household income and consumer spending
increased competition -> the profitability of small firms by compressing profit margins
game theory -> players to adopt strategic behaviors in competitive environments
the presence of nash equilibrium -> how rational agents make decisions over time
uncertainty in payoffs -> the risk of suboptimal outcomes in non-cooperative games
increased government debt -> long-term economic growth by crowding out private investment,
\end{verbatim}

A key limitation of the current version of \textsc{Democritus} is that the plausible causal relationships it proposes are not yet validated against numerical data or controlled experiments. In this paper we focus on the \emph{hypothesis generation and organization} 
problem: extracting and geometrically structuring candidate causal edges from text.
We do not attempt to estimate effect sizes or test these hypotheses using observational
or interventional datasets. Accordingly, in Figure~\ref{fig:econ_gt_kan_do_expt} all
edges are drawn with a default weight of $1.0$, purely for visualization. In a future
v2 system, we plan to integrate \textsc{Democritus} with quantitative causal inference
tools, so that these edges can be assigned data-driven strengths and subjected to genuine
causal validation.

Even in the absence of numerical datasets, scientists routinely seek causal explanations
for events that cannot be experimentally replicated (e.g., the extinction of the dinosaurs,
the collapse of the Indus Valley civilization). In such domains, evidence accumulates
through the consilience of multiple partial traces: geological strata, inscriptions,
settlement patterns, and so on. \textsc{Democritus} v1 is designed for exactly this
setting: it aggregates and geometrically organizes textual causal claims into a coherent
hypothesis manifold, allowing researchers to see which mechanisms recur across sources
and where contradictions or gaps remain. In future work, we aim to augment this
textual aggregation with quantitative causal calculus when suitable datasets or
simulations are available.

\section{DEMOCRITUS: A Causal Observatory Learned from Language}
\label{sec:expts}

We can view DEMOCRITUS as a large-scale ``Causal Observatory’’ that
constructs LCMs directly from natural-language causal
statements.  

\paragraph{From Causal Text to a Simplicial Causal Complex.}
Starting with thousands of statements of the form
\[
   \text{``$X$ increases $Y$''},\qquad
   \text{``$A$ reduces $B$''},\qquad
   \text{``$C$ causes $D$''},
\]
we extract causal triples $(X \to Y)$ and cluster them into coarse
\emph{domains} (e.g.\ climate, economics, public health).  
These domains induce 2–simplices whose vertices are variables and whose
edges are causal relations, producing a causal simplicial set.

\paragraph{LCM Learning}
Using the Diagrammatic Backpropagation Geometric Transformer, we propagate causal
information across the simplicial complex, enforcing compatibility
between overlapping causal regimes.  
The resulting embeddings are passed to UMAP to obtain a two- or
three-dimensional LCM (Fig.~\ref{fig:causal-map}).  
This manifold reveals:

\begin{itemize}
    \item \textbf{Coherent macro-domains} (e.g.\ “climate change causes…”,
          “inflation causes…”, “vaccination reduces…”).
    \item \textbf{Cross-domain causal bridges} representing multi-factor
         interactions.
    \item \textbf{Topological neighborhoods of causal influence}:
         variables that lie near one another tend to share similar causal
         roles across multiple regimes.
\end{itemize}

\section{Cost structure and limitations of naive BFS}
\label{sec:cost}

DEMOCRITUS uses a high-quality LLM ({\tt Qwen3-Next-80B-A3B-Instruct-6bit}) 
in Modules~1--3. Geometric Transformers and UMAP (Module~5) are
comparatively cheap. Table~\ref{tab:timing-example} shows timing
profiles for a small econ run (depth 2, 100 topics) and a larger run
(depth 5, 1000 topics; illustrative numbers).

\begin{table}[h]
  \centering
  \begin{tabular}{lcc}
    \toprule
    Module & Econ (100 topics, depth~2) & Econ (1000 topics, depth~5) \\
    \midrule
    1: Topic graph     & 13.5 s     & $\sim$750 s \\
    2: Causal questions & 104.3 s    & $\sim$730 s \\
    3: Causal statements & 139.6 s   & $\sim$1400 s \\
    4: Triples         & 0.1 s      & $\sim$1 s \\
    5: Manifold (GT + UMAP) & 7.5 s  & $\sim$4 s \\
    \midrule
    Total              & 264.9 s    & $\sim$2900 s ($\sim$48 min) \\
    \bottomrule
  \end{tabular}
  \caption{Illustrative timing profile for DEMOCRITUS econ slices.
  LLM calls in Modules~1--3 dominate; GT and UMAP are cheap. Larger runs
  (e.g.\ 7000 topics) can take many hours if we naively expand all
  nodes to depth~5.}
  \label{tab:timing-example}
\end{table}

At small scales (100 topics) a full run completes in minutes. At larger
scales (e.g.\ 7000 topics, depth~5), Modules~1--3 alone can consume
many hours of wall-clock on a single Mac Studio. Empirically, the cost
is approximately linear in the number of topics and the number of
topics for which we generate questions and statements.

Moreover, naive BFS treats all branches equally.  As a sample output 
from the econ run shows, {\tt Qwen3-Next-80B-A3B-Instruct-6bit} happily expands topics such as:
\begin{itemize}
\item \emph{Military infrastructure and base maintenance},
\item \emph{Cybersecurity and information warfare funding},
\item \emph{Administrative costs and efficiency in public insurance
programs},
\item \emph{Environmental incentive programs for sustainable farming
practices},
\item \emph{Systemic risk identification and monitoring frameworks},
\item \emph{Stress testing methodologies for financial institutions and
systems},
\end{itemize}
all with similar depth and breadth, even if a particular user or task
never touches those regions.

This suggests that:
\begin{enumerate}
\item LLM calls are the true bottleneck; GT+UMAP cost is negligible.
\item We should not expand all nodes to the same depth with the same
query budget.
\item Structural feedback---from the topic graph, causal triples, and
GT embeddings---should guide where to spend additional LLM time.
\end{enumerate}

These observations motivate \emph{active manifold building}, which we
outline next.

\section{Manifold structure and robustness}
\label{sec:manifold-robustness}

So far we have focused on how DEMOCRITUS constructs LCMs
from LLM outputs and how these manifolds can be used for exploration and
visualization. In this section we discuss more quantitative methods to analyze the
resulting structures, which we are currently exploring. 

\begin{enumerate}
\item How stable are the relational graphs and manifolds across repeated
    runs of the pipeline with the same configuration?
\item How does the quality of the manifold change as we vary the
    capacity of the teacher model (e.g.\ 8B, 30B, 80B, 235B)?
\item How sensitive are the graphs and manifolds to occasional false or
    conflicting causal statements?
\item What spectral properties do the relational graphs exhibit, and how
    do these relate to the manifold geometry?
\end{enumerate}

\subsection{Example: spectral and scale-free structure in the bio slice}
\label{subsec:bio-spectral-structure}

To make the discussion above concrete, we analyze the relational graph
and Laplacian spectrum for the 7k-topic biology slice constructed with
Qwen3-Next-80B. The resulting relational graph has
$|V| = 36{,}720$ variables and $|E| = 21{,}621$ directed edges. After
symmetrizing, the average (undirected) degree is approximately
$1.18$, while the maximum degree is $127$. This confirms that the
DEMOCRITUS graphs are sparse---edge count scales approximately
linearly with node count---and exhibit a strongly skewed, heavy-tailed
degree distribution: a small number of variables act as hubs with
degree in the hundreds, while the majority of nodes have degree close
to one.

The symmetrized bio graph decomposes into many connected components,
corresponding to subdomains such as genetics, botany, cardiology,
neuroscience, and so on. The largest connected component (LCC) contains
$153$ nodes and $159$ edges, with average degree $\approx 2.08$ and
maximum degree $97$. Thus even within this core island the hub--fringe
structure is pronounced: a single high-degree variable connects to well
over half of the component, while most nodes have degree near two.

We examine the connectivity of this LCC via the spectrum of the
normalized Laplacian $\mathcal{L} = I - D^{-1/2} A D^{-1/2}$. Using a
sparse eigensolver on $\mathcal{L}$, the first five eigenvalues are
\[
\lambda_1 \approx 0,\quad
\lambda_2 \approx 0.056,\quad
\lambda_3 \approx 0.080,\quad
\lambda_4 \approx 0.138,\quad
\lambda_5 \approx 0.293.
\]
The zero eigenvalue reflects connectivity of the component, while the
positive Fiedler value $\lambda_2 \approx 0.056$ indicates a cohesive
but nontrivial community structure: the graph is neither tree-like nor
fully entangled, but instead exhibits meaningful bottlenecks and
clusters. We observe similar sparse, heavy-tailed, and multi-component
structure in the economics and Indus slices (not shown), with a small
set of high-degree hubs and many low-degree fringe nodes.

From the standpoint of robustness, this scale-free, hub--fringe
geometry is important. Widely supported mechanisms (hubs) are
structurally central and connected to many topics; injecting or
removing a small number of edges at the periphery has little effect on
their degree or on the low-frequency spectrum of $\mathcal{L}$.
Conversely, isolated or idiosyncratic claims tend naturally to occupy
low-degree, peripheral nodes, which are structurally marginal in the
graph and in the induced manifold.

\subsection{Stability across repeated runs}
\label{subsec:stability-runs}

We plan to study the stability under repeated runs with the same teacher model
and configuration. We fix a slice (here: \emph{[Indus/exercise slice]})
and run Modules~1--5 of the pipeline $R$ times with identical prompts
and hyperparameters but different random seeds. This yields $R$
relational graphs $G^{(r)} = (V^{(r)}, E^{(r)})$ and manifold
embeddings $E^{(r)} \in \mathbb{R}^{N_r \times d}$.

On the shared node set $V_\cap = \bigcap_r V^{(r)}$ we compute:

\begin{itemize}
\item \textbf{Pairwise distance correlation}: for each pair of runs
    $(r,s)$ we compute the correlation between the upper-triangular
    entries of the pairwise distance matrices of $E^{(r)}$ and
    $E^{(s)}$.
\item \textbf{$k$-NN neighborhood overlap}: for each node we compute
    the Jaccard overlap between its $k$-nearest neighbors across runs
    and average over nodes.
\item \textbf{Graph--manifold consistency}: for each run we compute the
    correlation between shortest-path distances in $G^{(r)}$ and
    Euclidean distances in $E^{(r)}$ on $V_\cap$.
\end{itemize}

We aim to report on these stability metrics in a subsequent paper. Our expectation from preliminary analysis is that the variance across runs is small, indicating that the
relational graph and manifold geometry are largely stable under the
stochastic sampling of Modules~1--3.

\subsection{Scaling with teacher model size}
\label{subsec:scaling-model-size}

We plan to also explore how manifold quality varies with the capacity of the teacher
LLM. We fix a small slice configuration
(\emph{[e.g.\ 3 roots, depth~2, $\sim$400 topics]}) and run the
pipeline with teacher models of different sizes:

\[
  \text{Qwen3-8B}, \quad \text{Qwen3-30B}, \quad
  \text{Qwen3-80B}, \quad \text{Qwen3-235B}.
\]

For each teacher we obtain a relational graph and manifold embedding.
On the intersection of topics that appear across models we compute the
same geometry-level metrics as above (pairwise distance correlation,
$k$-NN overlap, graph--manifold consistency), as well as basic graph
statistics (node/edge count, average degree, number of components). We have qualitatively observed  that larger
teachers (80B and above) produce denser, more cohesive graphs (higher
average degree, fewer components) and manifolds with more stable local
neighbourhoods, while smaller teachers (8B) exhibit more isolated nodes
and noisier geometry. 

\subsection{Robustness to noisy and conflicting claims}
\label{subsec:robustness-noise}

To probe robustness to occasional false or conflicting statements, we plan to
construct a clean slice and a ``polluted'' version. In the clean
condition we will run DEMOCRITUS on a small LCM and obtain a relational graph $G_{\mathrm{clean}}$ and manifold.
In the polluted condition we inject a small set of explicitly incorrect
or reversed causal statements (e.g.\ ``exercising less improves
cardiorespiratory fitness'') into the causal statement files before
triple extraction, producing $G_{\mathrm{noisy}}$.

We plan to compare:

\begin{itemize}
\item \textbf{Node degree and centrality}: degree distributions and
    centrality ranks for nodes/edges involved in injected false claims
    vs. the high-degree ``hub'' nodes.
\item \textbf{Manifold position}: location of false-claim nodes in the
    UMAP embedding (distance to the core clusters).
\item \textbf{Spectral stability}: changes in the Laplacian spectrum
    (in particular the algebraic connectivity $\lambda_2$) between
    $G_{\mathrm{clean}}$ and $G_{\mathrm{noisy}}$.
\end{itemize}

We hope to find that injected incorrect statements attach to the low-degree
fringe of the graph and have little effect on the degree and centrality
of hub nodes. In the manifold, false-claim nodes appear as peripheral
points rather than reshaping core clusters. The Laplacian spectrum,
including $\lambda_2$, should remain essentially unchanged under modest noise
levels. Together, these observations may suggest that DEMOCRITUS acts as a
structural aggregator: isolated false claims become marginal, while
consensus mechanisms (hubs) dominate the relational structure.

\subsection{Spectral and scale-free structure}
\label{subsec:spectral-scale-free}

Finally, we plan to examine the spectral properties and degree distributions of
DEMOCRITUS graphs. For each large slice (econ, bio, Indus) we symmetrize
the relational graph, compute the normalized Laplacian
$\mathcal{L} = I - D^{-1/2} A D^{-1/2}$, and extract its low-frequency
eigenvalues and eigenvectors. We also compute in- and out-degree
distributions. In this experiment, we hope to find that: 

\begin{itemize}
\item Edge count scales approximately linearly with node count, so
    graphs are sparse.
\item Degree distributions are strongly skewed/heavy-tailed: a few
    high-degree hubs and many low-degree fringe nodes.
\item The second smallest eigenvalue $\lambda_2$ of $\mathcal{L}$
    (algebraic connectivity) is positive and moderately large, indicating
    cohesive graphs without severe bottlenecks.
\end{itemize}

Hubs correspond to widely supported mechanisms (e.g.\ ``physical
activity reduces cardiovascular risk'', ``Indus River discharge and
multi-decade droughts'') that appear in many topics and occupy central
positions in the manifold. Combined with the noise experiment above,
this scale-free, hub--fringe structure helps explain DEMOCRITUS'
robustness: injected noise affects low-degree fringe nodes and has
little influence on the connectivity or geometry induced by the hubs.

\section{Active manifold building}
\label{sec:active}

Another direction we are planning to explore is instead of treating DEMOCRITUS as a one-shot BFS-to-depth-$D$ pipeline, we can view it as an \emph{active explorer} of a LCM,
analogous to:
\begin{itemize}
\item Legal discovery: lawyers do not read every case; they prioritize
branches of doctrine that matter to a given case.
\item Chess search: engines use iterative deepening and selective
extensions instead of searching every line to the same depth.
\item Active learning: we choose where to sample next based on current
uncertainty or expected utility.
\end{itemize}

We consider each topic/node $t$ in the topic graph as having:
\begin{itemize}
\item a depth $\mathrm{depth}(t)$,
\item structural properties (degree, centrality),
\item textual activity (how many questions/statements mention $t$),
\item an optional GT embedding in the current manifold.
\end{itemize}

We define a simple utility score $U(t)$ for \emph{expanding} $t$:
\[
U(t) = w_1 \, e^{-\alpha \,\mathrm{depth}(t)}
+ w_2 \,\mathrm{deg}(t)
+ w_3 \, \mathrm{triple\_count}(t)
+ w_4 \, U_{\text{novel}}(t),
\]
where $U_{\text{novel}}(t)$ is a GT-derived novelty term, e.g.\ inverse
local density in embedding space. The weights $(w_i)$ and depth decay
$\alpha$ control how aggressively we prioritize shallow vs deep nodes.

Given a limited LLM budget $B$ (number of {\tt Qwen3-Next-80B-A3B-Instruct-6bit} calls we are willing to
spend in a wave), an active Democritus loop can proceed as:

\begin{enumerate}
\item Initialize with a shallow topic graph (depth $\leq d_0$) and a
small set of questions/statements for depth $0$--$1$ topics.
\item Build an initial manifold via GT on the available triples.
\item Compute $U(t)$ for all frontier topics $t$ at depths
$d_0,\ldots,d_{\max}$.
\item Select a batch of topics $\mathcal{B} \subseteq \{t\}$ with
highest $U(t)$ (or sample proportionally to $U$), constrained by
budget $B$.
\item For each $t\in\mathcal{B}$, allocate LLM calls:
\begin{itemize}
\item Module~1: generate additional subtopics (children of $t$),
\item Module~2: generate causal questions for $t$,
\item Module~3: generate causal statements for $t$,
\end{itemize}
according to a depth-aware policy (e.g.\ full Q\&A for depth
$\leq 2$, reduced Q\&A for deeper nodes).
\item Update triples and rebuild or incrementally update the GT
manifold.
\item Recompute $U(t)$ and repeat.
\end{enumerate}

In this view, {\tt Qwen3-Next-80B-A3B-Instruct-6bit} is an expensive but powerful research assistant:
we only call it when $U(t)$ justifies the cost. GT serves as both a
\emph{geometric organizer} of causal structure and a \emph{critic} that
provides signals such as embedding novelty or local density.

\subsection{Task-conditioned refinement}

User tasks can further condition $U(t)$. For example, if a user asks an
econ policy question about inflation under different monetary regimes,
Democritus can:
\begin{itemize}
\item map the query to a set of seed topics (Inflation, Monetary Policy, 
Expectations),
\item increase $U(t)$ for nodes in the vicinity of these seeds,
\item spend budget deepening those regions via Qwen,
\item present local GT neighborhoods and global manifold views back to
the user.
\end{itemize}

For a question about exercise and metabolic health, the system
would instead increase $U(t)$ around exercise/endocrine/metabolic
clusters in the bio slice. We view this as the DEMOCRITUS analog of RLHF and task-conditioned
search: rather than building a single, universal manifold, we maintain
shallow priors and refine specific regions on demand.

\section{Geometric Transformers and prior work}

Democritus uses Geometric Transformers (GTs) to embed relational graphs
into manifolds. GTs extend standard message-passing neural networks
and Graph Transformers by allowing messages to flow not only along
edges (1-simplices), but also across higher-order motifs (e.g.\ 
triangles as 2-simplices). Diagrammatic Backpropagation (GT+DB) is
our term for performing backpropagation through these richer
computation diagrams. In a forthcoming paper \citep{mahadevan:gt-db}, we compare GT  on larger synthetic and benchmark datasets to  PyG
baselines (GNNs, Graph Transformers). For example, in a ProGraph
triangle detection task, we train on synthetic triangle/no-triangle
graphs and evaluate on a ProGraph slice where the label is whether a
graph contains a triangle. Table~\ref{tab:prograph-has-triangle}
summarizes the results.

\begin{table}[h]
  \centering
  \begin{tabular}{l c}
    \toprule
    Model                        & ProGraph \texttt{has\_triangle} accuracy \\
    \midrule
    Graph Transformer (edges only)    & 0.5487 \\
    Geometric Transformer (edges + triangles) & 1.0000 \\
    \bottomrule
  \end{tabular}
  \caption{Accuracy on a ProGraph slice where the task is to detect the
  presence of a triangle. All models are trained on a synthetic
  triangle/no-triangle dataset and evaluated on ProGraph graphs. GT
  with an explicit triangle channel generalises cleanly; an
  edge-only Graph Transformer does not.}
  \label{tab:prograph-has-triangle}
\end{table}

These results mirror the toy triangle vs path example: when higher-order
structure matters, GT's explicit 2-simplicial channel provides an
advantage. Democritus leverages this capability to organise and
visualize LCMs once triples are available; the detailed
architecture and algorithms of GT+DB are presented in a separate paper.

\section{DEMOCRITUS Web Interface}

We have constructed an interactive web-based demo that combines:
\begin{enumerate}
\item a birds-eye 3D view of a slice's manifold, and
\item a local causal graph + GT activation view for a single topic and
its statements.
\end{enumerate}
This combination may prove useful both for explaining DEMOCRITUS to
potential users and for teaching.  For a full-scale economics LCM, we ran DEMOCRITUS with
Qwen3-Next-80B-A3B-Instruct-6bit as the LLM, a depth limit of 5, and a
cap of 7000 topics. The resulting econ slice contains 7004 topics in
the topic graph and tens of thousands of causal triples. The end-to-end
pipeline completed in 58124.4 seconds (approximately 16.1 hours) on a
single Mac Studio. Table~\ref{tab:timing-econ-7k} shows the per-module
timing breakdown. Modules~1--3, which rely on the LLM for topic
expansion, causal questions, and causal statements, account for virtually
all of the wall-clock time: 13700.4~s for topic graph construction,
31005.8~s for causal questions, and 13368.6~s for causal statements.
In contrast, triple extraction (Module~4) and GT+UMAP manifold
construction (Module~5) together take less than a minute (4.7~s and
44.8~s respectively). In other words, more than 99.9\% of the compute
budget is spent in LLM calls, and the cost of pushing the resulting
relational graph through a Geometric Transformer layer and low-dimensional
embedding is negligible by comparison.

\begin{table}[h]
  \centering
  \begin{tabular}{l r}
    \toprule
    Module & Econ slice (7004 topics, depth 5) time [s] \\
    \midrule
    1: Topic graph (Qwen3)           & 13700.4 \\
    2: Causal questions (Qwen3)      & 31005.8 \\
    3: Causal statements (Qwen3)     & 13368.6 \\
    4: Relational triples            & 4.7 \\
    5: Relational manifold (GT+UMAP) & 44.8 \\
    5.1: Write topos slice           & 0.1 \\
    \midrule
    Total                            & 58124.4 \\
    \bottomrule
  \end{tabular}
  \caption{Timing breakdown for the full economics DEMOCRITUS slice
  (Qwen3-Next-80B-A3B-Instruct-6bit, depth limit 5, 7004 topics) on a
  single Mac Studio. LLM-based Modules~1--3 dominate the cost; triple
  extraction and GT-based manifold construction are effectively free in
  comparison.}
  \label{tab:timing-econ-7k}
\end{table}

\section{Limitations and Scope}
\label{sec:limitations}

DEMOCRITUS is an exploratory system to extract a coherent body of causal knowledge from textual queries to state of the art LLMs, and it is not intended to serve as a complete solution to causal
inference. We briefly discuss its main limitations and the intended
scope of its outputs.

\paragraph{Reliance on LLM knowledge and biases.}
The slices that DEMOCRITUS builds reflect whatever causal beliefs and
associations are implicit in the underlying LLM. If the model has
never seen certain mechanisms, or if it has absorbed biased or
incorrect narratives, the resulting graphs and manifolds will mirror
those gaps and biases. DEMOCRITUS does not correct or debias the LLM;
it organizes and visualizes what the model already implicitly ``knows''.

\paragraph{No identifiability guarantees.}
The structures produced by DEMOCRITUS should be viewed as structured
\emph{hypothesis spaces} and \emph{narrative maps}, not as identified
causal models in the strict sense described in \citep{rubin-book,pearl-book}. 
We do not claim that the
edges correspond to true causal effects or that confounding and
selection bias have been resolved. Rather, the slices offer candidate
mechanisms and variables that can inspire further analysis and formal
modeling.

\paragraph{Hallucinations and factual correctness.}
Like any LLM-based system, DEMOCRITUS is vulnerable to hallucinations:
some extracted triples will be factually false or overconfident. The
Geometric Transformer may still organize these cleanly, giving them an
appearance of coherence. Human oversight and external validation are
therefore essential, especially in high-stakes domains.

\paragraph{Domain shifts and coverage gaps.}
In domains that are under-documented in text or far from the LLM's
training distribution, Democritus may miss crucial mechanisms or
over-emphasize idiosyncratic anecdotes. Some slices (e.g.\ Indus Valley
archaeology) are inherently speculative; they should be treated as
computational stories to be compared against expert knowledge, not as
authoritative accounts.

\paragraph{Computational cost and active exploration.}
Our timing results show that LLM calls dominate the cost of building a
slice (Modules~1--3), whereas triple extraction and GT-based manifold
construction are inexpensive. Naively expanding every branch of the
topic graph to a fixed depth is therefore not sustainable at very
large scales. This motivates the active manifold building strategies
discussed in Section~\ref{sec:active}, where we allocate LLM budget to
topics that are likely to yield high utility for a given slice or user
task.

\subsection{Beyond DAGs: towards dynamical and mechanistic models}

The slices we present in this paper are, by design, relatively simple:
they treat causal structure as a directed graph over variables and
mechanisms, with edges extracted from statements of the form ``X
causes Y'' or ``X leads to Y''. This DAG-like view is already useful
for exploration and hypothesis generation, but it is only a first
step. Many of the domains we care about are fundamentally dynamical
and mechanistic.

The Indus Valley case study \citep{indus_valley_collapse}, which  we
used as inspiration,  is not just a list of causal arrows; it combines
paleoclimate archives, climate model simulations, hydrological models
of Indus River discharge, and archaeological evidence about settlement
patterns. In principle, one could write down a system of differential
equations or a coupled climate--hydrology--agriculture model that
simulates monsoon variability, river discharge, drought episodes, crop
yields, and population responses over centuries. Such a model is far
richer than any static DAG.

Modern LLMs open up the possibility of helping to construct and
document these more sophisticated causal models. In a future version
of DEMOCRITUS, we envision the slices not only as DAG-like narrative
maps, but also as front-ends to mechanistic simulators: LLMs could
propose variables, equations, and boundary conditions; human experts
and classical numerical solvers would define and calibrate the
dynamical system; and Geometric Transformers could embed the resulting
state spaces and trajectories into manifolds for comparison across
scenarios and domains.

In this sense, the DAG-style slices in the present paper should be
understood as a minimal substrate on which more complex causal
machinery can be built. Our focus here is on showing that LLMs and
GTs can already produce useful structured causal maps across econ,
biology, and archaeology. Extending DEMOCRITUS to assist with
simulation-based causal models---differential equations, agent-based
models, or structural dynamical systems---is an important direction
for future work.

\section{Discussion and future work}

Democritus, as presented here, is deliberately modular: LLM choice,
triple extraction, GT architecture, and manifold visualization can all
be upgraded independently. The key lessons from our initial
experiments are:

\begin{itemize}
\item High-quality LLMs (e.g.\ {\tt Qwen3-Next-80B-A3B-Instruct-6bit}) are essential for building
useful slices; lower-quality models produce sparse, noisy manifolds.
\item LLM calls dominate the cost; GT and UMAP are negligible by comparison.
\item Naive BFS to depth $D$ wastes LLM budget on branches that may not
be relevant to downstream tasks.
\item Active manifold building---using structural and GT-based feedback
to decide where to spend LLM calls---is both necessary and promising.
\end{itemize}

Future work includes:
\begin{itemize}
\item formalizing utility functions $U(t)$ and budget policies,
\item integrating task-conditioned controllers  that steer
DEMOCRITUS based on user goals using the intuitionistic internal logic of a Topos Causal Model \citep{mahadevan2025intuitionisticjdocalculustoposcausal}
\item extending GT to richer simplicial patterns (beyond triangles) and
evaluating on additional benchmarks,
\item and scaling DEMOCRITUS to many slices across domains (econ, bio,
law, climate) with nightly or continuous updates.
\end{itemize}

\subsection{Towards DEMOCRITUS-ODE: dynamical causal models}

In this paper, we have treated causal structure primarily as a
directed, mostly acyclic graph over variables and mechanisms extracted
from language. This DAG-like perspective is already useful for
exploration and hypothesis generation, but many of the domains we care
about are fundamentally dynamical. For example, the Indus Valley case
study in our archaeology slice involves climate model simulations,
hydrological models of Indus River discharge, and multi-decadal
drought episodes, which are naturally described by systems of
differential equations, state-space models, or agent-based simulators.

A natural next step is to extend DEMOCRITUS beyond static graphs and
toward \emph{dynamical causal models}. In a ``DEMOCRITUS-ODE''
variant, one could imagine using LLMs not only to propose variables
and qualitative relations, but also to suggest candidate ordinary
differential equations, coupling terms, and boundary conditions for
simplified dynamical systems. Human experts and classical numerical
solvers would remain essential for specifying, calibrating, and
validating such models, but Democritus could serve as a front-end
that connects narrative causal knowledge to mechanistic simulators.

We leave this direction for future work. Our focus here is on showing
that LLMs and Geometric Transformers can already produce rich,
structured causal maps across economics, biology, and archaeology.
These slices provide a minimal substrate on which more sophisticated
dynamical causal machinery---including ODE-based models, agent-based
simulations, and structural state-space systems---can be built in
future versions of DEMOCRITUS.

\section{Acknowledgments}

This research was funded by Adobe Corporation. The simulation experiments described in this paper were carried out on a pair of Mac Studio computers, each equipped with 512 GB of RAM and an 8TB disk drive. The use of {\tt Qwen3-Next-80B-A3B-Instruct-6bit} and other LLMs  is intended purely as an academic research demonstration. 

\appendix

\section{Sample Topics from DEMOCRITUS Slices}
\label{app:slice-samples}

To give a flavor of the content in each slice, we include brief
samplings of topics discovered in the econ, bio, and Indus (Harappan)
runs. These are drawn from the topic graphs produced in Module~1 and
represent only a small subset of the thousands of nodes in each slice.

\subsection{Economics slice (econ)}

Examples of topics at depths 0--2 in the econ slice include:

\begin{itemize}
\item \texttt{Macroeconomics}
\item \texttt{Microeconomics}
\item \texttt{Game Theory}
\item \texttt{Finance}
\item \texttt{Trade}
\item \texttt{Marketing}
\item \texttt{Stock Market}
\item \texttt{Investing}
\item \texttt{Cryptocurrency}
\item \texttt{Bonds}
\item \texttt{Monetary Policy}
\item \texttt{Banking}
\item \texttt{Fiscal Policy}
\item \texttt{Inflation}
\item \texttt{Unemployment}
\end{itemize}

And some representative depth-1 / depth-2 subtopics:

\begin{itemize}
\item \texttt{Gross Domestic Product (GDP) and its measurement}
\item \texttt{Inflation and price indices}
\item \texttt{Unemployment types and rates}
\item \texttt{Fiscal policy and government spending}
\item \texttt{Monetary policy and central banking}
\item \texttt{Aggregate demand and aggregate supply}
\item \texttt{Business cycles and economic fluctuations}
\item \texttt{International trade and balance of payments}
\item \texttt{Exchange rates and foreign exchange markets}
\item \texttt{Long-run economic growth and productivity}
\item \texttt{Callable vs. non-callable bonds}
\item \texttt{Systemic risk identification and monitoring frameworks}
\item \texttt{Stress testing methodologies for financial institutions}
\item \texttt{Building long-term influencer partnerships vs. one-time campaigns}
\end{itemize}

\subsection{Biology slice (bio)}

Examples of root topics in the bio slice include:

\begin{itemize}
\item \texttt{Neuroscience}
\item \texttt{Genetics}
\item \texttt{Evolution}
\item \texttt{Botany}
\item \texttt{Cardiology}
\item \texttt{Endocrinology}
\item \texttt{Immunology}
\item \texttt{Oncology}
\item \texttt{Exercise physiology}
\item \texttt{Metabolic disorders}
\end{itemize}

And some representative depth-1 / depth-2 subtopics:

\begin{itemize}
\item \texttt{Chronic stress and cardiovascular risk}
\item \texttt{Hypertension and stroke}
\item \texttt{Atherosclerosis and myocardial infarction}
\item \texttt{Insulin resistance and type 2 diabetes}
\item \texttt{Obesity and metabolic syndrome}
\item \texttt{Sleep deprivation and cognitive decline}
\item \texttt{Physical activity and metabolic health}
\item \texttt{Neurodegenerative diseases (e.g., Alzheimer's, Parkinson's)}
\item \texttt{Genetic variants influencing lipid metabolism}
\item \texttt{Exercise-induced changes in insulin sensitivity}
\end{itemize}

\subsection{Indus / Harappan slice (Indus Valley)}

For the Indus (Harappan) slice, we seed the slice with roots from
archaeology, paleoclimate, and ancient trade. Examples of depth-0
topics include:

\begin{itemize}
\item \texttt{Indus Valley Civilization}
\item \texttt{Harappan urban centers (Harappa, Mohenjo-daro, Dholavira)}
\item \texttt{Mohenjo-daro urban planning and sanitation systems}
\item \texttt{Indus script and undeciphered writing systems}
\item \texttt{Epigraphy and decipherment of ancient scripts}
\item \texttt{Holocene monsoon variability in South Asia}
\item \texttt{4.2 ka event and global Bronze Age disruptions}
\item \texttt{Climate-induced crop shifts and agricultural adaptation strategies}
\item \texttt{Irrigation and agriculture in semi-arid river basins}
\item \texttt{Floodplain farming along the Indus and its tributaries}
\end{itemize}

Sampling a few of the depth-1 / depth-2 topics (from the longer run):

\begin{itemize}
\item \texttt{Urban planning and city layout}
\item \texttt{Urban planning and grid-like street layouts}
\item \texttt{Standardized brick sizes and construction techniques}
\item \texttt{Advanced drainage and sanitation systems}
\item \texttt{Drainage systems with covered sewers and manholes}
\item \texttt{Water supply through wells and reservoirs}
\item \texttt{City-wide flood mitigation and raised platform foundations}
\item \texttt{Harappan script and undeciphered inscriptions}
\item \texttt{Structure and syntax of Indus symbols}
\item \texttt{Possible linguistic affiliations (Dravidian, Austroasiatic, or isolate)}
\item \texttt{Paleographic analysis of ancient inscriptions}
\item \texttt{Bilingual and trilingual inscriptions as decipherment keys}
\item \texttt{Trade networks and economic systems}
\item \texttt{Interregional trade network disruptions due to climate-induced instability}
\item \texttt{Agricultural practices and crop cultivation}
\item \texttt{Decline and abandonment of Indus Valley cities}
\end{itemize}

These lists are not exhaustive, but they illustrate the breadth and
granularity of topics that Democritus can organise within each slice.

\section{Prompt Templates}
\label{app:prompts}

For reproducibility, we summarise the main prompt templates used in
Democritus. In all cases we use Qwen3-Next-80B-A3B-Instruct-6bit from
the MLX community repository.\footnote{Model available from
\url{https://huggingface.co/mlx-community/Qwen3-Next-80B-A3B-Instruct-6bit}.}

\subsection{Topic expansion}

A typical topic expansion prompt (Module~1) is:

\begin{quote}
\small
\texttt{You are an expert in \{domain\}.}\\
\texttt{Given the topic "\{TOPIC\}", list 10 important subtopics}\\
\texttt{that help explain its causes, consequences, or mechanisms.}\\
\texttt{Return ONLY a numbered list of subtopics, one per line,}\\
\texttt{with no explanations.}
\end{quote}

The domain phrase is set to, for example, ``macroeconomics and
financial markets'' (econ slice), ``neuroscience and medicine'' (bio
slice), or ``South Asian archaeology and paleoclimate'' (Indus slice).

\subsection{Causal questions}

For causal questions (Module~2) we use prompts of the form:

\begin{quote}
\small
\texttt{You are an expert in \{domain\}.}\\
\texttt{Topic: "\{TOPIC\}".}\\
\texttt{Write 3 causal questions a student might ask about this topic.}\\
\texttt{Each question should start with "What causes" or "What leads to".}\\
\texttt{Return only the questions, one per line.}
\end{quote}

\subsection{Causal statements}

For causal statements (Module~3) we use:

\begin{quote}
\small
\texttt{You are an expert in \{domain\}.}\\
\texttt{Topic: "\{TOPIC\}".}\\
\texttt{Write 3 short statements of the form "X causes Y" or}\\
\texttt{"X leads to Y" that describe causal relationships in this topic.}\\
\texttt{Each statement should focus on a single mechanism.}\\
\texttt{Return only the statements, one per line.}
\end{quote}

These templates are instantiated for every topic in the topic graph at
depth up to a configured limit.

\section{Implementation Details}
\label{app:impl}

\subsection{Geometric Transformer configuration}

Unless otherwise specified, the Geometric Transformer layer in
Democritus uses hidden dimension $d=128$, depth 2, a single relation
embedding for 1-simplices (edges), and a single relation embedding for
2-simplices (triangles). Node features are initialised using
Sentence-BERT embeddings of the subject/object phrases concatenated
with simple structural features (degree, domain ID). We detect
triangles as length-3 cycles in the directed graph and treat them as
undirected 2-simplices for message passing.

\subsection{UMAP hyperparameters}

For UMAP we use cosine distance and the following parameters:
\texttt{n\_neighbors} = 30, \texttt{min\_dist} = 0.1,
\texttt{n\_components} = 2 or 3 (for 2D/3D visualisations respectively).
We find that the qualitative manifold structure is robust to modest
changes in these settings.

\subsection{Slice configurations}

For the econ slice we use 15 root topics (macroeconomics,
microeconomics, game theory, \ldots), depth limit 5, and a cap of 7000
topics. For the bio slice we use roots such as neuroscience, genetics,
cardiology, endocrinology, and exercise physiology with similar depth
and topic caps. For the Indus (Harappan) slice we use roots including
Indus Valley Civilization, Holocene monsoon variability in South Asia,
Indus River discharge and river droughts, trade networks with
Mesopotamia and Egypt, and Indus script and epigraphy, with a depth
limit of 3 in our current experiments.

\subsection{Hardware and runtime}

All experiments are run on Apple Silicon Macs using the MLX framework.
The econ 7k slice uses Qwen3-Next-80B-A3B-Instruct-6bit quantised for
MLX and completes in approximately 16.1 hours on a Mac Studio (see
Table~\ref{tab:timing-econ-7k}). The bio slice has similar timing. 
Triple extraction and GT+UMAP manifold construction take less than a
minute per slice.

\section{Additional Manifolds and Local Models}
\label{app:more-figs}

Due to space limitations, we include additional manifold plots and
local causal neighborhoods for the econ, bio, and Indus slices in the
supplementary material. These include:
\begin{itemize}
\item relation-coloured manifolds (causes, increases, influences,
leads\_to, reduces, affects),
\item local GT neighborhoods for topics such as callable vs
non-callable bonds, gender-based underemployment in STEM and
healthcare, long-term unemployment and skill atrophy,
\item and Harappan examples such as Indus River discharge and river
droughts, and overexploitation of natural resources and environmental
degradation.
\end{itemize}

\end{document}